\RequirePackage{fix-cm}
\documentclass[twocolumn]{svjour3}          % twocolumn
\smartqed  % flush right qed marks, e.g. at end of proof
\usepackage{graphicx}
\usepackage{natbib}
\usepackage{multirow}
\usepackage{amsmath}
\usepackage{amssymb}
\usepackage{bbm}
\usepackage{bm}
\usepackage{times}
\usepackage{booktabs}
\usepackage{array} % for extended column definitions, e.g. >{\em}c
\usepackage{blindtext}
\usepackage{comment}
\usepackage[breaklinks=true]{hyperref}
\usepackage{caption, subcaption}
\usepackage{stfloats}
\usepackage{makecell}
\hypersetup{
    colorlinks=true,
    linkcolor=blue,
    filecolor=blue,      
    urlcolor=blue,
    citecolor=blue,
}
\usepackage{pifont}

\usepackage[dvipsnames]{xcolor}
\usepackage{color, colortbl}
\definecolor{citecolor}{HTML}{0071bc}
\definecolor{tabhighlight}{HTML}{e5e5e5}
\makeatletter
\renewcommand\paragraph{
  \@startsection{paragraph} % name
  {4} % level
  {\z@} % indent
  {.5em \@plus1ex \@minus.2ex} % beforeskip
  {-.5em} % afterskip
  {\normalfont\normalsize\bfseries} % style
}
\makeatother
\begin{document}
\sloppy

\title{Ultra-High-Deﬁnition Image Restoration: New Benchmarks and A Dual Interaction Prior-Driven Solution %\thanks{Grants or other notes
}
% \subtitle{Do you have a subtitle?\\ If so, write it here}

\author{Liyan Wang$^{1}$ \and
        Cong Wang$^{2}$ \and
        Jinshan Pan$^{3}$ \and
        Xiaofeng Liu$^{4}$ \and
        \\
        Weixiang Zhou$^{1}$ \and
        Xiaoran Sun$^{1}$ \and
        Wei Wang$^{5}$ \and
        Zhixun Su$^{1,6}$
}
% \author {Han Liang\Mark {1} \hfill \and \hfill Wenqian Zhang\Mark {2} \hfill \and \hfill Wenxuan Li\Mark {1} \hfill \and \hfill Jingyi Yu\Mark {1} \hfill \and \hfill Lan Xu\Mark {1}}

% \authorrunning{Short form of author list} % if too long for running head

% Communicated by Ling Shao,Hubert P. H. Shum, Timothy Hospedales.
% Anurag Ranjan and David T. Hoffmann have contributed equally to this work.
% \institute{xxx and xxx have contributed equally to this work}

\institute{Liyan Wang \at
              \email{wangliyan@mail.dlut.edu.cn}
           \and
           Cong Wang \at
              \email{supercong94@gmail.com}
           \and
           Jinshan Pan \at
              \email{sdluran@gmail.com}
            \and
            Xiaofeng Liu \at
            \email{xiaofeng.liu@yale.edu}
           \and
          Weixiang Zhou \at
              \email{s20201162006@mail.dlut.edu.cn}
           \and
           Xiaoran Sun \at
              \email{sunxiaoran@mail.dlut.edu.cn}
           \and
          Wei Wang \at
              \email{wangwei29@mail.sysu.edu.cn}
           \and
           Zhixun Su \at
              \email{zxsu@dlut.edu.cn}   
           \\
           $^1$ Dalian University of Technology, China
           \\
           $^2$ Centre for Advances in Reliability and Safety, Hong Kong
           \\
           $^3$ Nanjing University of Science and Technology, China 
           \\
           $^4$ Yale University, USA 
           \\
           $^5$ Shenzhen Campus of Sun Yat-Sen University, China 
           \\ 
           $^6$ Key Laboratory for Computational Mathematics and Data Intelligence of Liaoning Province, China 
}
\date{Received: date / Accepted: date}
\maketitle
\begin{abstract} \label{abstract}
Ultra-High-Definition (UHD) image restoration has acquired remarkable attention due to its practical demand.
In this paper, we construct UHD snow and rain benchmarks, named UHD-Snow and UHD-Rain, to remedy the deficiency in this field.
The UHD-Snow/UHD-Rain is established by simulating the physics process of rain/snow into consideration and each benchmark contains 3200 degraded/clear image pairs of 4K resolution.
Furthermore, we propose an effective UHD image restoration solution by considering gradient and normal priors in model design thanks to these priors' spatial and detail contributions.
Specifically, our method contains two branches: (a) feature fusion and reconstruction branch in high-resolution space and (b) prior feature interaction branch in low-resolution space.
The former learns high-resolution features and fuses prior-guided low-resolution features to reconstruct clear images, while the latter utilizes normal and gradient priors to mine useful spatial features and detail features to guide high-resolution recovery better.
To better utilize these priors, we introduce single prior feature interaction and dual prior feature interaction, where the former respectively fuses normal and gradient priors with high-resolution features to enhance prior ones, while the latter calculates the similarity between enhanced prior ones and further exploits dual guided filtering to boost the feature interaction of dual priors.
We conduct experiments on both new and existing public datasets and demonstrate the state-of-the-art performance of our method on UHD image low-light enhancement, dehazing, deblurring, desonwing, and deraining.
The source codes and benchmarks are available at \url{https://github.com/wlydlut/UHDDIP}.


\end{abstract}

\section{Introduction}\label{Introduction}
Recently, as imaging and acquisition equipment developed by leaps and bounds, Ultra-High-Definition (UHD) images with high pixel density and high resolution (4K images containing 3840 $\times$ 2160 pixels) have continuously acquired attention.
Compared with general images, UHD images naturally present more details and a wider color gamut, leaving continuous improvement of people's quality requirements for UHD images, which requires a systematic set of benchmark studies including the construction of relevant datasets and the design of algorithms suitable for processing UHD restoration.
Since current general learning-based image restoration algorithms~\citep{mm20_wang_dcsfn,mm20_wang_jdnet,wang2021single,yao2021pyramid,zhao_lie,chen2021hdcwnet,ChenFDTK20jstasr,ZamirA0HK0021mprnet,Zamir2021Restormer,0002LZCC21hinet,TuTZYMBL22maxim,wang2024selfpromer,wang2022online,wang2024promptrestorer,wang2024msgnn,MeiFZYZLFHS23,ZhangLCGBS22} cannot effectively process UHD images~\citep{LiGZLZFL23,wang2024uhdformer}, several UHD restoration models are developed~\citep{ZhengRCHWSJ21,DengRYWSC21,LLformer,LiGZLZFL23,wang2024uhdformer} as well as UHD restoration benchmarks. 

\begin{figure*}[t]
\centering
\begin{center}
\begin{tabular}{cccccc}
\hspace{-2mm}\includegraphics[width=\linewidth]{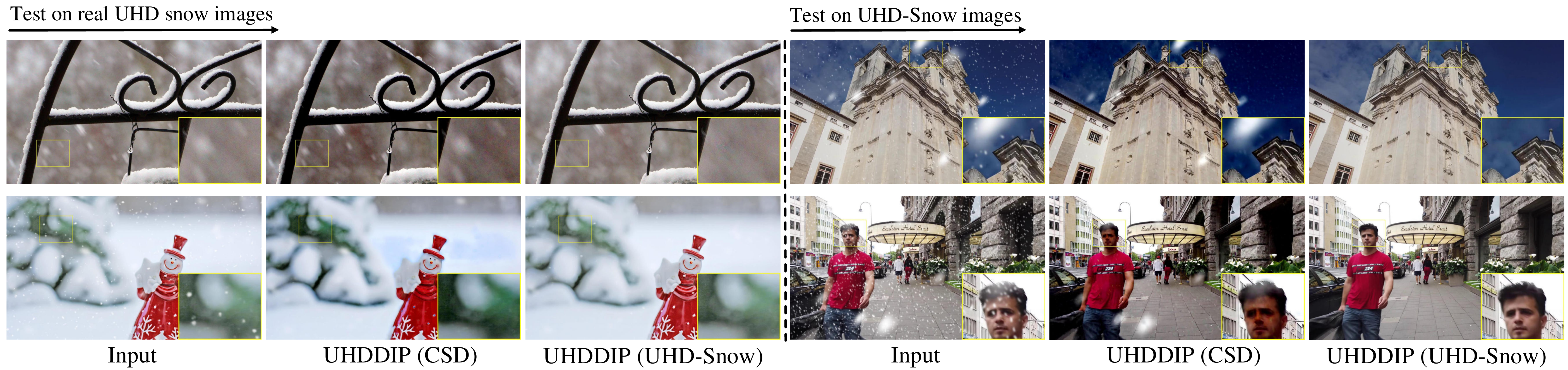} 
\end{tabular}
\caption{The UHDDIP model, when trained on low-resolution desnowing datasets CSD~\citep{ChenFHTCDK21CSD}, fails to remove larger snowflakes and preserve the original image's colors, while its performance remarkably improves after training on the proposed UHD-Snow dataset. Moreover, the visual results tested on the real UHD images (left three columns) and the proposed synthesized UHD-Snow images (right three columns) are consistent.}\label{fig: differences of low-res and high-res}
\end{center}
\end{figure*}

The specialized datasets for snow and rain UHD scenarios are largely unavailable at present, which hinders further exploration and research on the related tasks.
If models are trained solely on previously synthesized low-resolution datasets, they exhibit significant shortcomings when applied to UHD images. For example, as illustrated in Fig.~\ref{fig: differences of low-res and high-res} (second column), training on the low-resolution desnowing dataset CSD~\citep{ChenFHTCDK21CSD} results in difficulties removing larger snowflakes in real-world UHD snow images. Importantly, it tends to over-enhance, causing color inconsistencies with the original image. In contrast, training on the UHD dataset not only effectively removes snowflakes but also preserves the original image's colors (Fig.~\ref{fig: differences of low-res and high-res}, third column).
Hence, constructing UHD snow and rain benchmarks is practically needed.

Recently, various approaches have been proposed for UHD restoration including multi-guided bilateral learning for UHD image dehazing~\citep{ZhengRCHWSJ21}, multi-scale separable-patch integration networks for video deblurring ~\citep{DengRYWSC21}, Transformer-based LLFormer~\citep{LLformer}, Fourier embedding network UHDFour~\citep{LiGZLZFL23} for UHD low-light image enhancement, and UHDformer~\citep{wang2024uhdformer} by exploring the feature transformation from high- and low-resolution for UHD image restoration.
However, these methods predominantly focus on the input image’s intrinsic features, not taking into account additional relevant prior information conditions in model design, which makes it challenging to fully recover UHD images with more details and texture.

\begin{figure*}[t]
\centering
\includegraphics[width=1.0\textwidth]{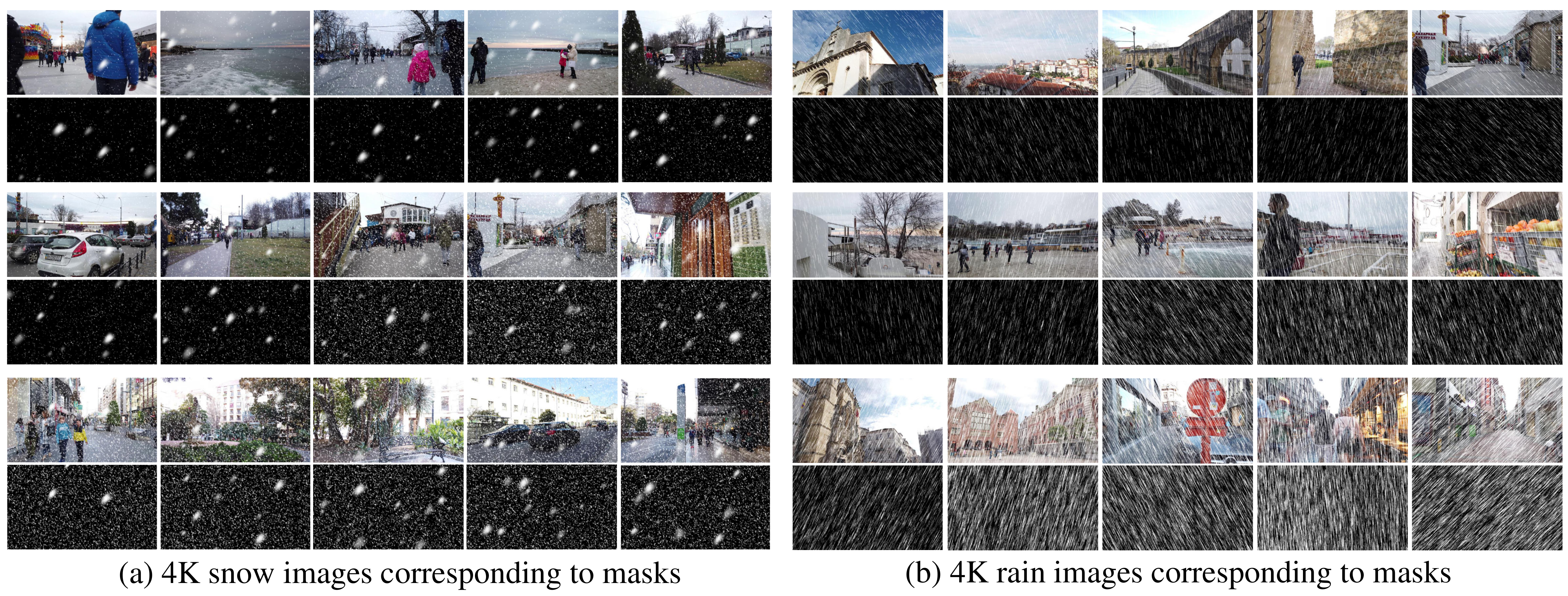}
\caption{The rain/snow images and their corresponding masks sampled from the proposed UHD-Snow and UHD-Rain datasets. Each dataset includes 3200 degraded/clean image pairs with 4K resolution (3000 pairs for training and 200 pairs for testing), which are synthesized snowflakes, snow streaks, and rain streaks with different densities, orientations, and locations.
}\label{fig:UHD datasets}
\end{figure*}

In this paper, we first construct two new benchmarks, named UHD-Snow and UHD-Rain, to facilitate the research of UHD image desnowing and deraining. Each dataset includes 3200 clean 4K images and corresponding degraded images synthesized by snowflakes and rain streaks with different densities, orientations, and locations, respectively.
Among them, 3000 pairs are used for training and 200 pairs for testing. 
Fig.~\ref{fig:UHD datasets} shows some examples from UHD-Snow and UHD-Rain benchmarks.
Furthermore, we propose an effective dual interaction prior-driven UHD restoration solution (UHDDIP).
UHDDIP is built on two interesting observations: 1) the normal map contains shaped regions or boundaries of the texture that could provide more geometric spatial structures (e.g., \cite{sfpnet} utilize normal prior to mitigate texture interference.); 2) the gradient map reveals each local region's edge and texture orientation that could render detail compensation.
The interaction between normal and gradient priors allows for the synthesis of complementary information.
Integrating these priors into model design will facilitate UHD restoration with finer structures and details.
Based on these, we suggest the UHDDIP to solve UHD restoration problems effectively. 

Specifically, UHDDIP contains two branches: (a) feature fusion and reconstruction branch in high-resolution space and (b) prior feature interaction branch in low-resolution space.
The former learns high-resolution features and fuses prior-guided low-resolution features to reconstruct final clear images, while the latter explores prior feature interaction to render improved features with finer structures and detail features for high-resolution space.
To better fuse and interact with prior features, we propose the prior feature interaction, containing single prior feature interaction (SPFI) and dual prior feature interaction (DPFI). 
The SPFI respectively fuses normal prior and gradient prior with high-resolution features to enhance prior ones, while the DPFI calculates the similarity between enhanced prior features and further exploits dual guided filtering to boost dual prior feature interaction for capturing better structures and details.
Through interaction, these priors dynamically enhance each other's strengths, leading to more robust features for guiding the high-resolution restoration.
The main contributions are summarized as follows:

\begin{itemize}
    \item We construct UHD-Snow and UHD-Rain benchmarks, each containing 3200 degraded/clean 4K image pairs, the first benchmarks used for UHD image desnowing and UHD image deraining so far as we know.
    %\vspace{1mm}
    \item We propose a general UHD image restoration framework, UHDDIP, which is built around a prior feature interaction module containing SPFI and DPFI, that incorporates gradient and normal priors into model design to achieve high-quality restoration with finer structures and details.
    %\vspace{1mm}
    \item Experiments on existing UHD datasets and established UHD-Snow and UHD-Rain benchmarks show that our method outperforms the state-of-the-art approaches on UHD low-light image enhancement, dehazing, deblurring, desnowing, and deraining.
\end{itemize}
\section{Related Work}\label{Related Works}
%This section reviews general image restoration methods, existing UHD restoration benchmarks, and UHD image restoration approaches.
\vspace{-4mm}
\textbf{\subsection{General Image Restoration}}
General Image Restoration (IR) aims to recover a clean image from its degraded observation, a task crucial in various fields such as photography, medical imaging, and satellite imagery.
Traditionally, IR methods have relied heavily on hand-crafted priors, including sparse coding \citep{LuoXJ15}, self-similarity \citep{BuadesCM05}, and gradient prior \citep{XuZJ13}. However, they often falter when applied to real-world scenarios due to their limited robustness and generalization capabilities.
The advent of deep learning marks a significant shift in the field, enabling the development of image restoration methods with the powerful implicit learning ability from large-scale data. 
By building various deep neural networks such as Convolutional Neural Network (CNN)~\citep{AlexNet,he2016deep}, Residual Network (ResNet)~\citep{ZhangTKZF21}, UNet\citep{ZhangLZZGT22}, Transformer~\citep{vision_transformer}, and have made extraordinary progress in the areas of image super-resolution~\citep{DongLHT16}, image deraining~\citep{ChenPLFL23,HuangYCHT21,LiRWATJCWC21}, desnowing ~\citep{ChenFHTCDK21CSD,ChenFDTK20}, deblurring~\citep{10061365,wang2021uformer}, 
dehazing~\citep{DehazeFormer,YangWGT24}, denoising~\citep{ZhangZCM017, Zamir2021Restormer}, etc.
However, as models grow in complexity, there is a simultaneous increase in computational demands. \cite{ChenCZS22nafnet} propose NAFNet, a simple baseline that forgoes traditional activation functions and replaces the Transformer with plain block, achieving advanced performance with reduced complexity.
\vspace{-2mm}
\textbf{\subsection{UHD Restoration Benchmarks}}
There have been several recent efforts to produce UHD datasets for specific image restoration tasks. 
For example,
\cite{ZhangLLRS0L021} firstly provide two large-scale datasets, UHDSR4K and UHDSR8K collected from the Internet for image super-resolution. 
UHDSR4K includes 8099 images with size of 3,840 × 2,160, and UHDSR8K includes 2966 images with size of 7,680 × 4,320. 
\cite{ZhengRCHWSJ21} create a 4K image dehazing dataset 4KID, which consists of 10,000 frames of hazy/sharp images extracted from 100 video clips by several different mobile phones.
Based on UHDSR4K and UHDSR8K, \cite{LLformer} create the UHD-LOL by synthesizing corresponding low-light images.
To process real low-light images with noise, \cite{LiGZLZFL23} collect a real low-light image enhancement dataset UHD-LL, that contains 2,150 low-noise/normal-clear 4K image pairs captured in different scenarios.
In addition, \cite{DengRYWSC21} propose a 4KRD dataset comprised of blurry videos and corresponding sharp frames using three smartphones.
To our knowledge, there are currently no UHD benchmark datasets specifically designed for UHD image desnowing and deraining. To address this gap, we construct two new benchmarks dedicated to UHD snow and rain scenarios.
\vspace{-4mm}
\textbf{\subsection{UHD Image Restoration Approaches}}
Based on the above UHD benchmark datasets, a series of restoration works have made exciting progress in using deep learning-based approaches to recover clear UHD images. 
For example, \cite{ZhengRCHWSJ21} propose a multi-guided bilateral upsampling model for UHD image dehazing.
To address the UHD video deblurring problem, \cite{DengRYWSC21} develop a separable-patch integration network by working with a multi-scale integration scheme.
\cite{LLformer} propose LLFormer, a transformer-based low-light image enhancement method, which is built by the axis-based multi-head self-attention and cross-layer attention fusion block.
In contrast to the above approaches with UHD images in the spatial domain, \cite{LiGZLZFL23} combine the Fourier transform into low-light image enhancement by utilizing amplitude and phase in a cascade network.
However, the above methods usually rely on large-capacity models for optimal performance, and the recovered results are often unsatisfactory.
\cite{wang2024uhdformer} propose a small-capacity model, UHDformer, to solve the UHD image restoration problem, which demonstrates excellent performance on three tasks, by exploring the feature transformation from high- and low-resolution for UHD image restoration.
In contrast to existing methods that focus solely on extracting features from the input image itself and do not yet explore other related prior information, we propose a dual interaction prior-driven UHD image restoration framework, which innovatively integrates additional priori information and mines more robust features to facilitate restoration.
\vspace{-2mm}
\section{UHD-Snow and UHD-Rain Datasets}\label{UHD-snow and UHD-Rain Datasets}
We first introduce the source and construction of proposed UHD-Snow and UHD-Rain datasets in Sec.~\ref{Sec:Source and Construction}.
Next, the mask generation process for UHD rain and snow is provided in Sec.~\ref{Sec:Mask Generation}.
%We first introduce the constructed benchmarks and then present a novel UHD restoration solution.
%\textbf{\subsection{Benchmarks}}
%%%%%%%%%%%%%%%%%%%%%%%%%%%%%%%
\begin{figure}[b]
\vspace{-2mm}
\centering
\includegraphics[width=0.42\textwidth]{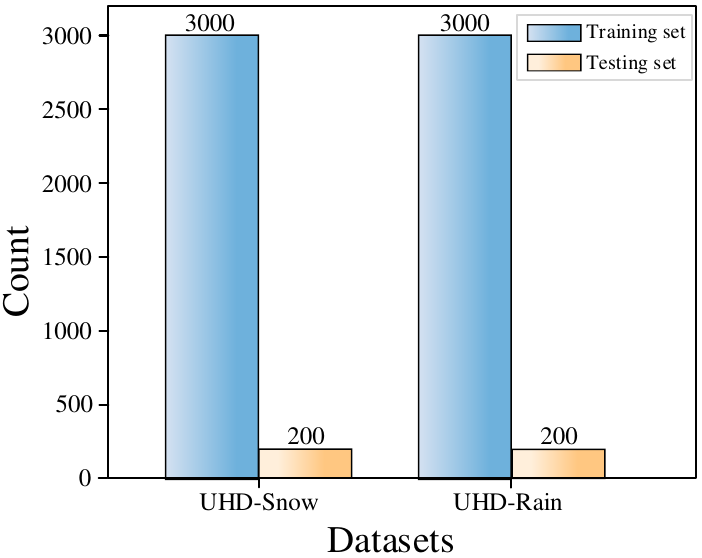}
% \vspace{-1mm}
\caption{Statistics of our constructed UHD-Snow and UHD-Rain benchmarks.}
\label{fig:datasets statistics}
\end{figure}
%%%%%%%%%%%%%%%%%%%%%%%%%%%%%%%
\vspace{-2mm}
\textbf{\subsection{Source and Construction}\label{Sec:Source and Construction}}
We construct two UHD-Snow and UHD-Rain datasets composed of 4K images of 3, 840 × 2, 160 resolution based on large-scale image datasets UHDSR4K~\citep{ZhangLLRS0L021}.
%%%
In detail, we extract 3200 original UHD images from the UHDSR4K dataset used for UHD image super-resolution to synthesize corresponding rain and snow images following Photoshop's rain and snow synthesis tutorial.
\begin{figure*}[t]
\centering
\includegraphics[width=0.9\textwidth]{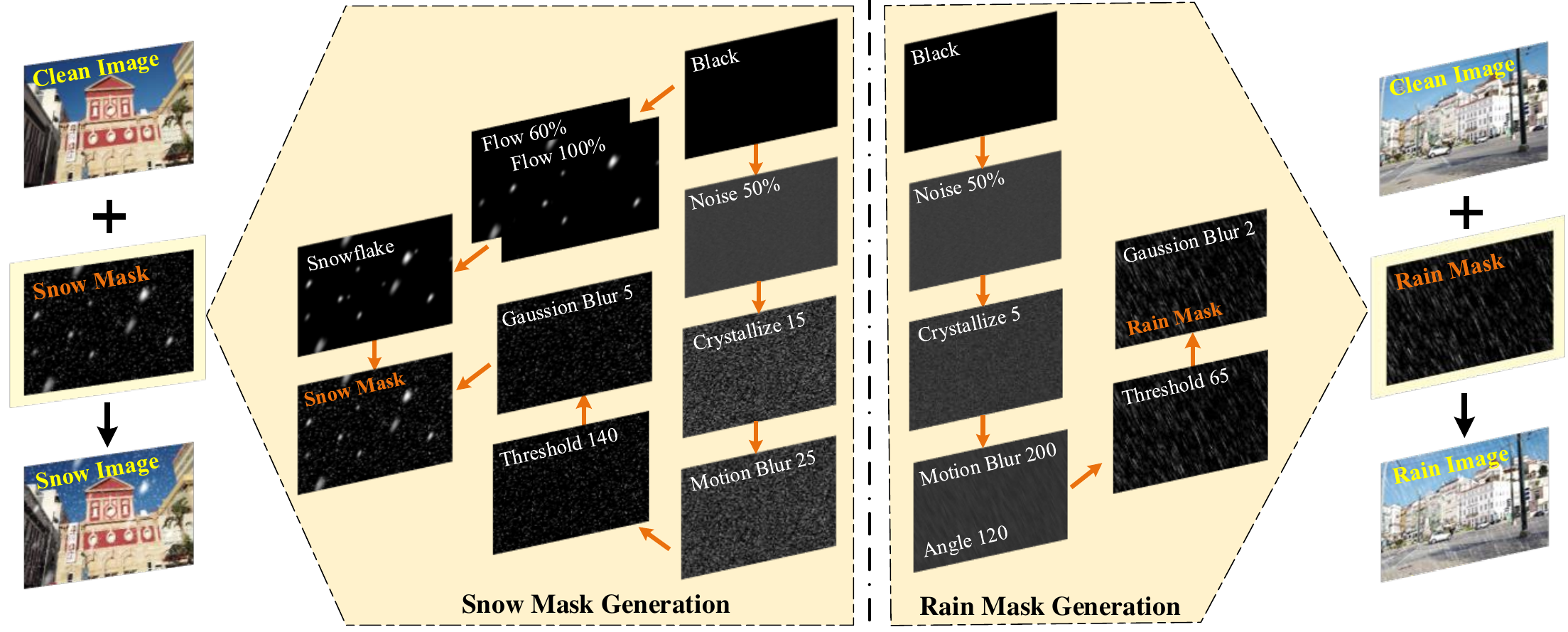}
\caption{The flowchart of snow (left) and rain (right) mask generation. 
In the snow mask generation process, we set the parameters as follows: $50\%$ noise (amount), $15$ crystallize (cell size), $25$ pixels motion blur (distance), a threshold level range of $100$-$165$ to control the density of the snow streaks, $5$ pixels Gaussian blur (radius), and two flows of $60\%$ and $100\%$ to create snowflakes with varying transparency.
For the generation of rain mask, the parameters are set: $50\%$ noise, $5$ crystallize, $200$ pixels motion blur with an angle range of $45^{\circ}$-$135^{\circ}$, a threshold level range of $55$-$67$, and $2$ pixels Gaussian blur.
}
\label{fig:masksnow_rain}
\end{figure*}
\begin{table*}[t]
\setlength{\tabcolsep}{11.3pt}
\caption{Photoshop's parameter settings for mask generation (50 different angles with a range of $45^{\circ}$-$135^{\circ}$ include $45^{\circ}, 50^{\circ}, 55^{\circ}, 60^{\circ}$-$80^{\circ}, 85^{\circ}, 95^{\circ}, 100^{\circ}$-$120^{\circ}, 125^{\circ}, 130^{\circ}, 135^{\circ}$).}\label{tab:Datasets Parameters}
\centering
\begin{tabular}{l|ccccccc}
\Xhline{1.5pt}
\textbf{Datasets} &\textbf{Noise} &  \textbf{Crystallize} &  \textbf{Motion Blur} &  \textbf{Gaussian Blur}   &\textbf{Threshold} &\textbf{Angle} &\textbf{Flow}
\\
\Xhline{1pt}
\textbf{UHD-Snow}&$50\%$ &$15$ &$25$ &$5$ &$100$-$165$  &- &$60\%$, $100\%$ \\
\textbf{UHD-Rain}&$50\%$ &$5$ &$200$ &$2$ 
&$55$-$67$  &$45^{\circ}$-$135^{\circ}$ &- \\ 
\Xhline{1.5pt}
\end{tabular}
\end{table*}
%%%%%%%%%%%%%%%%%%%%%%%%%%%%%%%
UHDSR4K provides 8099 images of 3, 840 × 2, 160 resolution collected from the Internet (Google, Youtube, and Instagram) containing diverse scenes such as city scenes, people, animals, buildings, cars, natural landscapes, and sculptures.
The training set is constructed by selecting the first 2,800 images from the training set and the first 200 images from the testing set in UHDSR4K (excluding indoor images), ensuring continuity with several adjacent images depicting the same scene. The testing set comprises 200 images randomly chosen from various scenes within the testing set of UHDSR4K, excluding the initial 200 images already utilized in the training set. 

Based on the USDSR4K dataset, we synthesize snowflakes and snow streaks with different sizes and densities following the snow mask of CSD~\citep{ChenFHTCDK21CSD} and rainy streaks with different densities and orientations following the rain mask of Rain100L~\citep{YangTFGYL20} and Rain100H~\citep{YangTFLGY17}.
Finally, the snow and rain masks are added to the corresponding UHD clean images to synthesize our UHD-Snow and UHD-Rain, which contain 3000 pairs for training and 200 pairs for testing, respectively.
The statistics of our UHD datasets are summarized in Fig.~\ref{fig:datasets statistics}.
\vspace{-4mm}
\textbf{\subsection{Mask Generation}\label{Sec:Mask Generation}}
\noindent For the training set, we create 600 different UHD snow and rain masks. Among them, the snow masks contain snow streaks with 10 different densities and snowflakes with different sizes, and the rain masks contain rain streaks with 50 different orientations and 4 different densities, which ensures the diversity of the generation. Furthermore, we adopt the Gaussian blur on rain particles and snow particles to better simulate real-world rain and snow scenarios. We apply the same manner to the testing set to recreate 200 different UHD snow and rain masks. Note that UHD-Snow and UHD-Rain datasets share the same clean images, which ensures that differences in model performance are due to weather effects rather than scene variation.

In the synthesis process, we first randomly generate noise and then apply crystallization, motion blur, Gaussian blur, threshold adjustment, etc.
For the snow mask, we set the parameters as follows: $50\%$ noise (amount), $15$ crystallize (cell size), $25$ pixels motion blur (distance), $5$ pixels Gaussian blur (radius), and a threshold level range of $100$-$165$ to control the density of the snow streaks. Additionally, we adopt two flows of $60\%$ and $100\%$ to create snowflakes with varying transparency.
\begin{figure*}[t]
\centering
\includegraphics[width=0.98\textwidth]{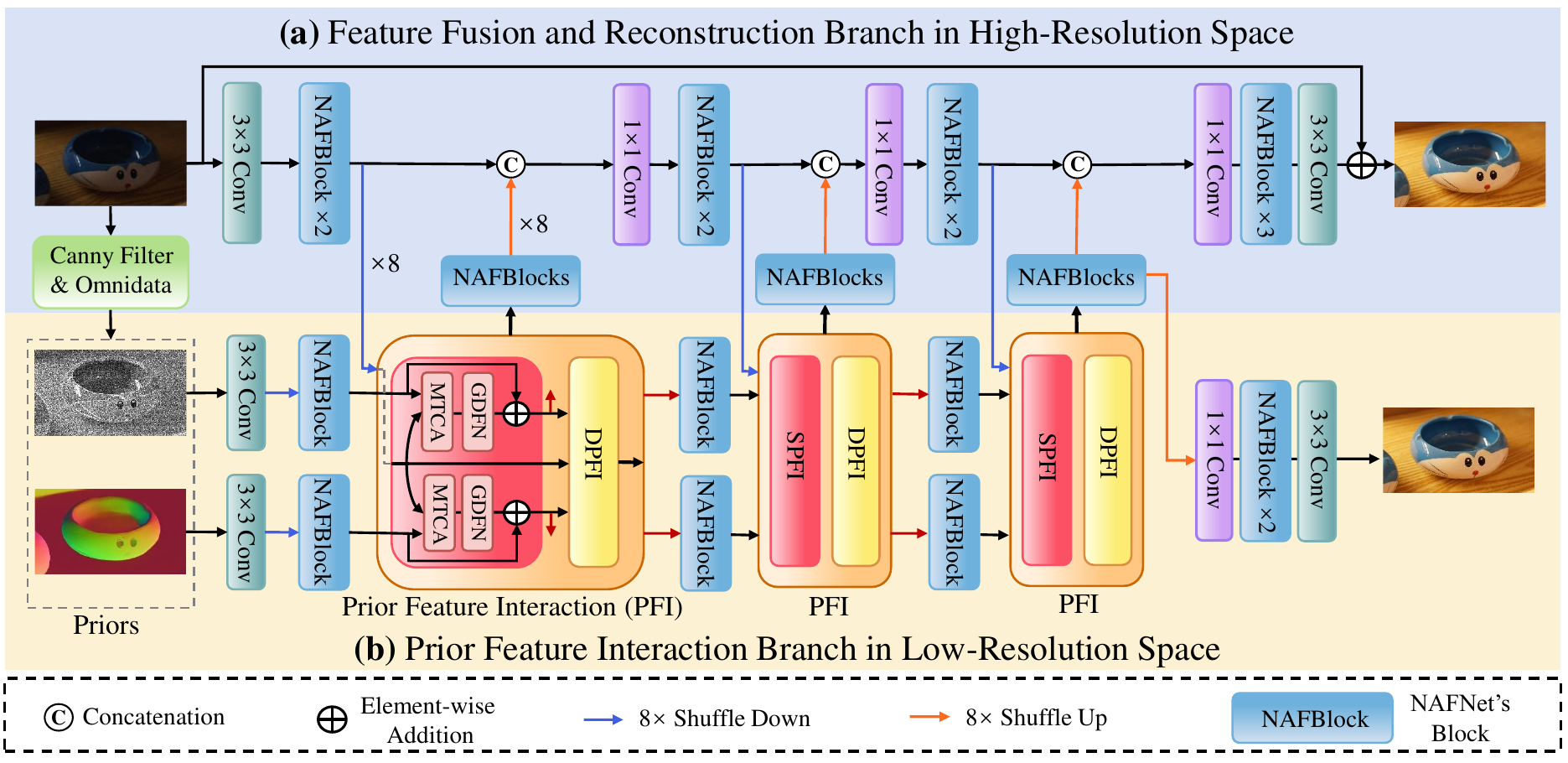} %fig1
\put(-462,201){\scriptsize{Input $\textbf{U}$}}
\put(-468,94){\scriptsize{Gradient $\textbf{P}\hspace{-0.5mm}_{g}$}}
\put(-465,52){\scriptsize{Normal $\textbf{P}\hspace{-0.5mm}_{n}$}}
\put(-402,190){\scriptsize{$\textbf{F}_{0}$}}
\put(-374,190){\scriptsize{$\textbf{F}_{1}$}}
\put(-324,135){\scriptsize{$\textbf{F}_{1}^{p}$}}
\put(-256,190){\scriptsize{$\textbf{F}_{2}$}}
\put(-179,190){\scriptsize{$\textbf{F}_{3}$}}
\put(-45,165){\scriptsize{Restored $\textbf{O}$}}
\put(-51,76){\scriptsize{Intermediate}}
\put(-45,68){\scriptsize{Result $\textbf{H}$}}
%\put(-308,85){\tiny{$\textbf{P}\hspace{-0.5mm}_{g\hspace{-0.3mm}1}$}}
\put(-285,119){\scriptsize{$\textbf{P}\hspace{-0.5mm}_{g\hspace{-0.3mm}1}^{'}$}}
\put(-285,76){\scriptsize{$\textbf{P}\hspace{-0.5mm}_{n\hspace{-0.5mm}1}^{'}$}}
\caption{Overview of our Dual Interaction Prior-driven UHD restoration network (UHDDIP).
UHDDIP contains two branches: (a) feature fusion and reconstruction branch in high-resolution space, which fuses low-resolution features into high-resolution space and reconstructs final images; (b) prior feature interaction branch in low-resolution space to modulate normal and gradient prior into useful features to guide high-resolution learning.
We utilize NAFBlock~\citep{ChenCZS22nafnet} as basic feature learning units.}
\label{fig:overall}
\end{figure*}
For the rain mask, the parameters are set as follows: $50\%$ noise, $5$ crystallize, $200$ pixels motion blur, $2$ pixels Gaussian blur, and a threshold level range of $55$-$67$ to control the density of the rain streaks. We then apply motion blur with an angle range of $45^{\circ}$-$135^{\circ}$.
Detailed parameter settings are shown in Table~\ref{tab:Datasets Parameters}.
The flowchart illustrating the mask generation for our UHD-Snow and UHD-Rain datasets is depicted in Fig.~\ref{fig:masksnow_rain}.
\vspace{4mm}
\section{Overview of UHDDIP}\label{Overview of UHDDIP}
\begin{figure*}[t]
\centering
\begin{center}
\begin{tabular}{c}
\includegraphics[width=1\linewidth]{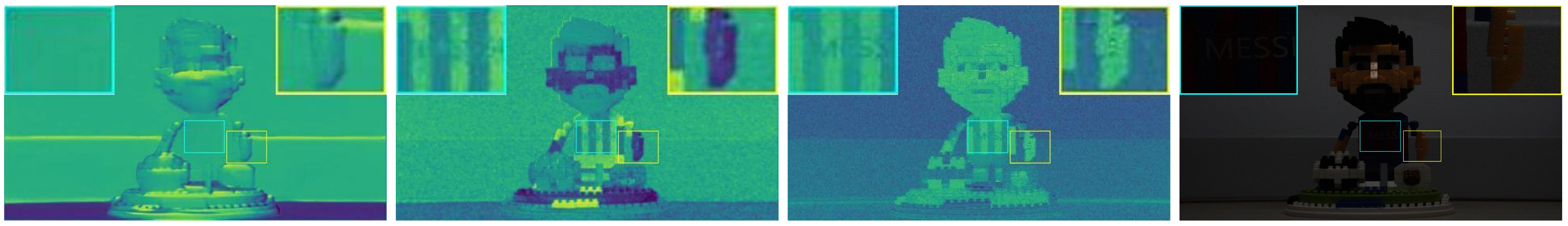} 
\\
 \makebox[0.245\textwidth][c]{(a) Normal $\textbf{P}_{n1}$ before PFI}\makebox[0.245\textwidth][c]{ (b) Normal $\textbf{P}_{n1}^{'}$ after SPFI}
 \makebox[0.245\textwidth][c]{(c) $\textbf{F}_{1}$ Before PFI }
 \makebox[0.245\textwidth][c]{(d) Input image}
\\
\includegraphics[width=1\linewidth]{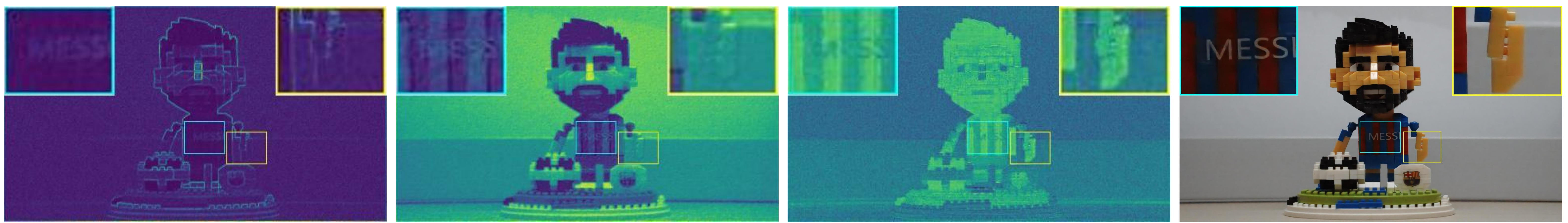} 
\\
 \makebox[0.245\textwidth][c]{(e) Gradient $\textbf{P}_{g1}$ before PFI}\makebox[0.245\textwidth][c]{ (f) Gradient $\textbf{P}_{g1}^{'}$ after SPFI}
 \makebox[0.245\textwidth][c]{ (g) $\textbf{F}_{1}^{p}$ after PFI }
 \makebox[0.245\textwidth][c]{(h) Restored image }
\\
\end{tabular}
\caption{Feature visualization at the first SPFI and DPFI module.
The normal prior feature (a) contains better structures, while the gradient prior feature (e) possesses richer details.
Compared to without prior features (c), using normal prior alone (b) helps to enhance the geometric structure of the image, like the arm’s boundaries. Applying only gradient prior (f) can improve finer texture details, like the letters' clarity. 
Whereas, utilizing in conjunction (g) performs the best.
Through interaction, two priors dynamically enhance each other’s strengths, to produce a result with finer structures and richer textures.
}
\label{fig:Feature visualization before and after the PFI}
\end{center}
\end{figure*}

Fig.~\ref{fig:overall} depicts our proposed UHDDIP, containing two branches: (a) feature fusion and reconstruction branch in high-resolution space, which fuses low-resolution features into the high-resolution space and reconstructs final latent clean images; (b) prior feature interaction branch in the low-resolution space to modulate normal and gradient priors into useful features to guide high-resolution learning.
\textbf{\subsection{Feature Fusion and Reconstruction Branch in High-Resolution Space}}
\noindent As depicted in Fig.~\ref{fig:overall}(a), given a degraded UHD image $\textbf{U}$~$\in$~$\mathbb{R}^{H\times W \times 3}$ as input, UHDDIP firstly applies a $3 \times 3$ convolution layer to extract shallow feature $\textbf{F}_0$~$\in$~$\mathbb{R}^{H\times W \times C}$, where $H\times W$ denotes the spatial dimension and $C$ is the number of channels. 
Next, $\textbf{F}_0$ is fed into the first group of NAFBlocks to obtain the first-level high-resolution feature $\textbf{F}_1$.
Then, $\textbf{F}_1$ is passed by $8 \times$ shuffled-down into the low-resolution space to produce $\textbf{F}_{1}^{p}$ that interacts with the prior features, and then $\textbf{F}_{1}^{p}$ is fed into NAFBlocks for further learning. 
The output features are $8 \times$ shuffled-up to be concatenated with the first-level high-resolution feature $\textbf{F}_1$ at channel dimension and a new first-level feature $\textbf{F}_1^{'}$ is obtained after a $1 \times 1$ convolution.
Then, the fused feature $\textbf{F}_1^{'}$ is fed into the second group NAFBlocks and the same action continues to be performed until after the third group NAFBlocks, high-resolution feature reconstruction starts to be implemented. 
Finally, the output generated after three NAFBlocks and a $3\times3$ convolutional layer is added to the input image to obtain the final restored image $\textbf{O}$~$\in$~$\mathbb{R}^{H\times W\times3}$.
\textbf{\subsection{Prior Feature Interaction Branch in Low-Resolution Space}}
\noindent To provide richer structures and details, we respectively operate on the input image $\textbf{U}$ using the Omnidata~\citep{EftekharSMZ21} and the Canny filter to generate normal prior $\textbf{P}_{n}$~$\in$~$\mathbb{R}^{H\times W \times 3}$ and gradient prior $\textbf{P}_{g}$~$\in$~$\mathbb{R}^{H\times W \times 1}$.
\begin{figure*}[t]
\centering
\includegraphics[width=1.0\textwidth]{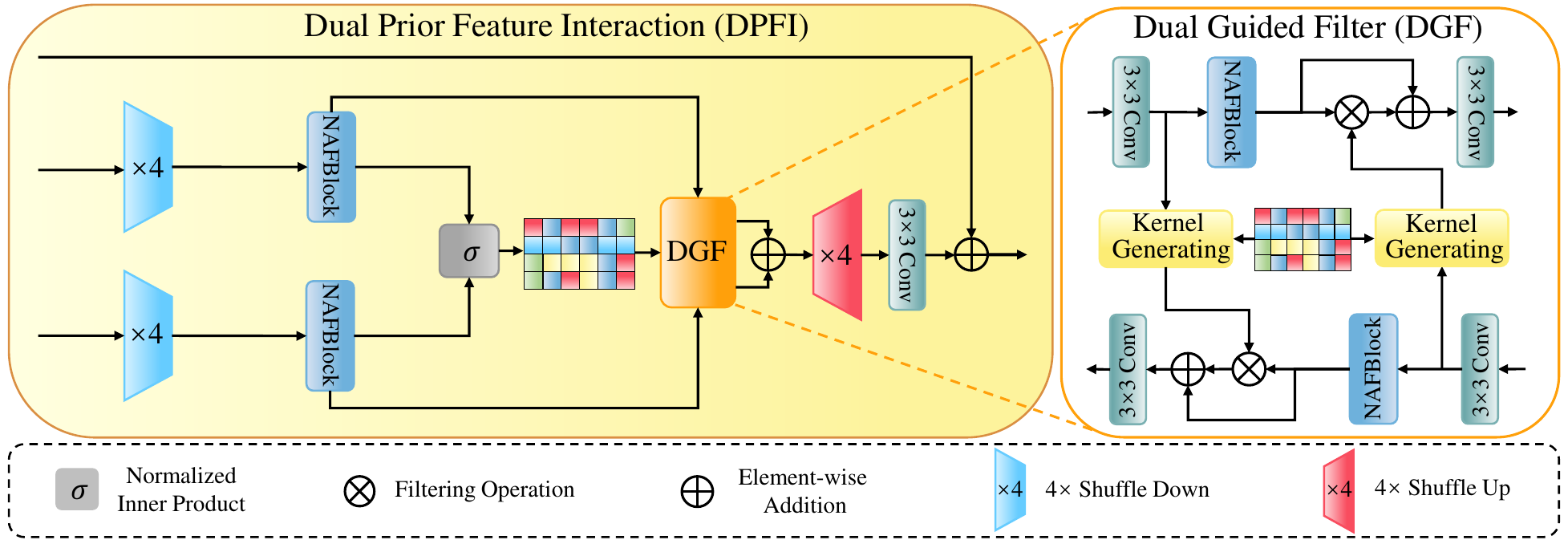}
\put(-489,145)
{\scriptsize{$C\hspace{-0.9mm}\times\hspace{-0.9mm}\tilde{H}\hspace{-0.9mm}\times\hspace{-0.9mm}\tilde{W}$}}
\put(-489,109)
{\scriptsize{$C\hspace{-0.9mm}\times\hspace{-0.9mm}\tilde{H}\hspace{-0.9mm}\times\hspace{-0.9mm}\tilde{W}$}}
\put(-489,57)
{\scriptsize{$C\hspace{-0.9mm}\times\hspace{-0.9mm}\tilde{H}\hspace{-0.9mm}\times\hspace{-0.9mm}\tilde{W}$}}
\put(-436,109)
{\scriptsize{$C\hspace{-0.9mm}\times\hspace{-0.9mm}\frac{\tilde{H}}{4}\hspace{-0.9mm}\times\hspace{-0.9mm}\frac{\tilde{W}}{4}$}}
\put(-436,56)
{\scriptsize{$C\hspace{-0.9mm}\times\hspace{-0.9mm}\frac{\tilde{H}}{4}\hspace{-0.9mm}\times\hspace{-0.9mm}\frac{\tilde{W}}{4}$}}
\put(-309,132)
{\scriptsize{$C\hspace{-0.9mm}\times\hspace{-0.9mm}\frac{\tilde{H}}{4}\hspace{-0.9mm}\times\hspace{-0.9mm}\frac{\tilde{W}}{4}$}}
\put(-309,47)
{\scriptsize{$C\hspace{-0.9mm}\times\hspace{-0.9mm}\frac{\tilde{H}}{4}\hspace{-0.9mm}\times\hspace{-0.9mm}\frac{\tilde{W}}{4}$}}
\put(-382,124)
{\scriptsize{$(C\hspace{-0.6mm}*\hspace{-0.6mm}k\hspace{-0.6mm}*\hspace{-0.6mm}k)\hspace{-0.9mm}\times\hspace{-0.9mm}\frac{\tilde{H}\tilde{W}}{16}$}}
\put(-382,56)
{\scriptsize{$(C\hspace{-0.6mm}*\hspace{-0.6mm}k\hspace{-0.6mm}*\hspace{-0.6mm}k)\hspace{-0.9mm}\times\hspace{-0.9mm}\frac{\tilde{H}\tilde{W}}{16}$}}
\put(-378,111){\footnotesize{unfold}}
\put(-378,68){\footnotesize{unfold}}
\put(-316,71){\small{$\textbf{W}$}}
\put(-324,108){\scriptsize{$1\hspace{-0.9mm}\times\hspace{-0.9mm}\frac{\tilde{H}\tilde{W}}{16}$}}
\put(-195,78)
{\scriptsize{$C\hspace{-0.9mm}\times\hspace{-0.9mm}\tilde{H}\hspace{-0.9mm}\times\hspace{-0.9mm}\tilde{W}$}}
\put(-89,108){\small{$\textbf{W}$}}
\put(-20,141){\scriptsize{$\textbf{F}\hspace{-0.2mm}_{p\hspace{-0.2mm}g}$}}
\put(-17,59){\scriptsize{$\textbf{F}\hspace{-0.2mm}_{n}$}}
\put(-157,60){\scriptsize{$\textbf{F}\hspace{-0.2mm}_{p\hspace{-0.2mm}n}$}}
\put(-155,140.5){\scriptsize{$\textbf{F}\hspace{-0.2mm}_{g}$}}
\put(-180,96){\scriptsize{$\textbf{F}\hspace{-0.2mm}^{p}$}}
\put(-482,122){\scriptsize{$\textbf{P}\hspace{-0.2mm}_{n}^{'}$}}
\put(-482,71){\scriptsize{$\textbf{P}\hspace{-0.2mm}_{g}^{'}$}}
\put(-482,156){\scriptsize{$\textbf{F}\hspace{-0.2mm}$}}
\put(-50,109){\scriptsize{$\textbf{K}\hspace{-0.2mm}_{n}$}}
\put(-125,80){\scriptsize{$\textbf{K}\hspace{-0.2mm}_{g}$}}
\caption{Dual Prior Feature Interaction (DPFI). DPFI aims to compute the similarity between two enhanced single prior features by employing their intrinsic properties to form interacted prior features, which provides meaningful guidance for feature learning in high-resolution space.}
\label{fig:DPFI}
\end{figure*}
As shown in Fig.~\ref{fig:overall}(b), in the low-resolution space, $\textbf{P}_{n}$ and $\textbf{P}_{g}$ are first fed into a ${3\times3}$ convolution and the first NAFBlock, to produce the first-level prior features $\textbf{P}_{n1}$ and $\textbf{P}_{g1}$, which serve as the input of the first prior feature interaction (PFI) module. 
At the same time, PFI also receives $\textbf{F}_{1}$ passed down from the high-resolution and encodes and interacts it along with $\textbf{P}_{n1}$ and $\textbf{P}_{g1}$ to generate interacted low-resolution feature $\textbf{F}_{1}^{p}$, and enhanced prior features $\textbf{P}_{n1}^{'}$, $\textbf{P}_{g1}^{'}$ that are fed into the second PFI to continue performing the same actions.
Until $\textbf{P}_{ni}^{'}$ and $\textbf{P}_{gi}^{'}$ and $\textbf{F}_{i}^{p}$ are obtained after the output of the $i^{\text{th}}$ ($i= 1, \dots, L$) PFI, these features are aggregated into the high-resolution branch to participate in the reconstruction of the final image. 
Moreover, the result $\textbf{H}$ generated after two NAFBlocks and a $3\times3$ convolutional layer is further used to supervise the low-resolution branch.
%\vspace{-2mm}
\textbf{\subsection{Prior Feature Interaction}}\label{PFI}
\noindent PFI contains two sub-modules: Single Prior Feature Interaction (SPFI) and Dual Prior Feature Interaction (DPFI).
The SPFI respectively fuses normal prior and gradient prior with high-resolution features to enhance prior ones, while the DPFI calculates the similarity between enhanced prior features and further exploits dual guided filtering to boost dual prior feature interaction.
\\
\\
\noindent\textbf{Single Prior Feature Interaction.} The main challenge of utilizing normal and gradient priors guides the image restoration lies in how to enable the network to be aware of image details and structure at the pixel level effectively.
To this end, we employ SPFI to enhance single prior features.
As shown in Fig.~\ref{fig:Feature visualization before and after the PFI}(a) and Fig.~\ref{fig:Feature visualization before and after the PFI}(e), we extract the features of the normal and gradient priors before entering the first SPFI, respectively.
Intuitively, normal features naturally provide the finer geometrical structure (arm's boundaries in the yellow box), while gradient features contain more texture details (letters in the blue box), and they can provide complementary information.
These features, respectively, are integrated with high-resolution features by the SPFI module to further enhance corresponding structures and details (See Fig.~\ref{fig:Feature visualization before and after the PFI}(b) and Fig.~\ref{fig:Feature visualization before and after the PFI}(f)).

Specifically, in SPFI, $\textbf{P}_{ni}$ and $\textbf{P}_{gi}$ are respectively fused with $\textbf{F}_{i}$ through two groups of Multi-Dconv Head Transposition Cross Attention (MTCA), which is defined as (For simplicity, we denote $\textbf{P}_{ni}$, $\textbf{P}_{gi}$, $\textbf{F}_{i}$ below as $\textbf{P}_{n}$, $\textbf{P}_{g}$, $\textbf{F}$):

% \begin{small} 
\begin{equation}
\begin{array}{ll}
\textit{MTCA}\Big(\textrm{$\textbf{Q}_{p}^{n}$},  \textrm{$\textbf{K}_{f}$},  \textrm{$\textbf{V}_{f}$}\Big)=
\textrm{Softmax\Big(\textrm{$\textbf{Q}_{p}^{n}$}\textrm{$\textbf{K}_{f}^{T}$} / \textrm{$\sqrt{f_k}$}}\Big)\textrm{$\textbf{V}_{f}$},
\end{array}
\label{eq:MTCA1}
\end{equation}
% \end{small}
%
% \begin{small} 
\begin{equation}
\begin{array}{ll}
\textit{MTCA}\Big(\textrm{$\textbf{Q}_{p}^{g}$},  \textrm{$\textbf{K}_{f}$},  \textrm{$\textbf{V}_{f}$}\Big)=
\textrm{Softmax\Big(\textrm{$\textbf{Q}_{p}^{g}$}\textrm{$\textbf{K}_{f}^{T}$} / \textrm{$\sqrt{f_k}$}}\Big)\textrm{$\textbf{V}_{f}$},
\end{array}
\label{eq:MTCA2}
\end{equation}
% \end{small}

\noindent where the query $\textbf{Q}_{p}^{n}$ is derived from normal prior feature $\textbf{P}_{n}$;
$\textbf{Q}_{p}^{g}$ is derived from grad prior feature $\textbf{P}_{g}$;
the key $\textbf{K}_{f}$ and value $\textbf{V}_{f}$ are derived from image feature $\textbf{F}$. These matrices are generated through layer normalization, $1\times1$ convolutions, and $3\times3$ depth-wise convolutions as orders.

Then, we employ the Gated-Dconv Feed-forward Network (GDFN)~\citep{Zamir2021Restormer} to generate single prior features $\textbf{P}_{n}^{'}$ and $\textbf{P}_{g}^{'}$ based on the attention map and the original prior features:

% \begin{small} 
\begin{equation}
\begin{array}{ll}
\textrm{$\textbf{P}_{n}^{'}$}
=\textit{GDFN}\Big(\textrm{$\textbf{P}_{n}$} + \textit{MTCA}\Big(\textrm{$\textbf{Q}_{p}^{n}$},  \textrm{$\textbf{K}_{f}$},  \textrm{$\textbf{V}_{f}$}\Big)\Big),
\end{array}
\label{eq:GDFN1}
\end{equation}
% \end{small} 
%
% \begin{small} 
\begin{equation}
\begin{array}{ll}
\textrm{$\textbf{P}_{g}^{'}$}
=\textit{GDFN}\Big(\textrm{$\textbf{P}_{g}$} + \textit{MTCA}\Big(\textrm{$\textbf{Q}_{p}^{g}$},  \textrm{$\textbf{K}_{f}$},  \textrm{$\textbf{V}_{f}$}\Big)\Big),
\end{array}
\label{eq:GDFN2}
\end{equation}
% \end{small} 

Finally, $\textbf{P}_{n}^{'}$ and $\textbf{P}_{g}^{'}$, as enhanced prior features of the current stage, are fed into the subsequent DPFI to implement the dual prior feature interaction and are also passed to the next SPFI to learn further.
\\
\\
\noindent\textbf{Dual Prior Feature Interaction.}
DPFI aims to compute the similarity between two enhanced single priors by employing their intrinsic properties to further capture image structures and details, which provides meaningful guidance for high-resolution space.
As shown in Fig.~\ref{fig:DPFI}, we first downsample features  $\textbf{P}_{n}^{'}$~$\in$~$\mathbb{R}^{C\times \tilde{H} \times \tilde{W}}(\tilde{H}=H/8, \tilde{W}=W/8)$ and $\textbf{P}_{g}^{'}$~$\in$~$\mathbb{R}^{C\times \tilde{H} \times \tilde{W}}$ by a factor of 4 to reduce the subsequent computational burden. 
They are then unfolded by a $k * k$ kernel ($k=3$) after passing through the NAFBlock in parallel, yielding patches of size $(C*k*k) \times \tilde{H}\tilde{W}/16$.
We compute all patches similarity using normalized inner product~\citep{LuLTLJ21}, and further obtain similarity weight $\textbf{W}$~$\in$~$[1 \times \tilde{H}\tilde{W}/16]$.
Motivated by~\citep{sfpnet}, we apply two prior features and similarity weight to the Dual Guided Filter (DGF) to further filter out irrelevant features while balancing the structure and detail. 

Specifically, DGF accepts two prior features $\textbf{F}_{n}$, $\textbf{F}_{g}$ and similarity weights $\textbf{W}$ as inputs and then generates the normal prior filter kernel and the gradient prior filter kernel, denoted as $\textbf{K}_{n}$, $\textbf{K}_{g}$:

% \begin{small}
\begin{equation}
\begin{array}{ll}
\textrm{$\textbf{K}_{n}$}
=\textit{g}\Big(\textit{Conv}_{3}(\textbf{F}_{n}), \textbf{W}\Big),
\textrm{$\textbf{K}_{g}$}
=\textit{g}\Big(\textit{Conv}_{3}(\textbf{F}_{g}), \textbf{W}\Big),
\end{array}
\label{eq:kernel}
\end{equation}
% \end{small}

\noindent where $\textit{g}$ denotes kernel generating module, containing a $1\times1$ convolution and an activation function. 
%$\textit{Conv}(\cdot)$ is a $3\times3$ convolution layer.
%

Then, the two kernels filter the two prior features separately to preserve their respective prior attributes and further filter out irrelevant features:

% \begin{small}
\begin{equation}
\begin{array}{ll}
\textrm{$\textbf{F}_{pn}$}
=\textbf{F}_{n} \bigotimes \textbf{K}_{g} + \textbf{F}_{n},
\quad
\textrm{$\textbf{F}_{pg}$}
=\textbf{F}_{g} \bigotimes \textbf{K}_{n} + \textbf{F}_{g},
\end{array}
\label{eq:kernel}
\end{equation}
% \end{small}

\noindent where $\bigotimes$ is the filtering operation; note that here $\textbf{F}_n$ and $\textbf{F}_{g}$ are obtained after a $3\times3$ convolution and a NAFBlock.
Finally, prior features are added to the low-resolution features passed down from the high-resolution to produce an interacted prior feature $\textbf{F}^{p}$.

Fig.~\ref{fig:Feature visualization before and after the PFI}(g) presents visualization features in the first PFI. 
One can observe that PFI can effectively improve both structure and detail features, which further facilitates restoration.
\noindent\textbf{\subsection{Loss Function}}
\noindent Following UHDformer~\citep{wang2024uhdformer}, we optimize our UHDDIP by minimizing the $L_1$ loss and frequency loss~\citep{0002JHJK21} between the restored result $\mathbf{O}$, intermediate result $\mathbf{H}$ and ground truth $\mathbf{G}$:

% \begin{small}
\begin{equation}
\begin{array}{ll}
\mathcal{L}_{\text{total}} &=  \|\mathbf{O}-\mathbf{G}\|_1 + \lambda \|\mathcal{F}(\mathbf{O})-\mathcal{F}(\mathbf{G})\|_1 \\\\
& + \, \alpha\|\mathbf{H}-\mathbf{G}\|_1 + \lambda \|\mathcal{F}(\mathbf{H})-\mathcal{F}(\mathbf{G})\|_1,
\end{array}
\label{eq:kernel}
\end{equation}
% \end{small}

\noindent where $\|\cdot\|_{1}$ denotes $L_1$ norm; $\mathcal{F}$ denotes the Fast Fourier transform; $\alpha$ and $\lambda$ are weights that are empirically set to 0.5 and 0.1.
%\vspace{4mm}
\section{Experiment}\label{Experiment}
\begin{figure}[t]
\centering
\begin{center}
\begin{tabular}{ccc}
\hspace{-4mm}\includegraphics[width=0.49\textwidth]{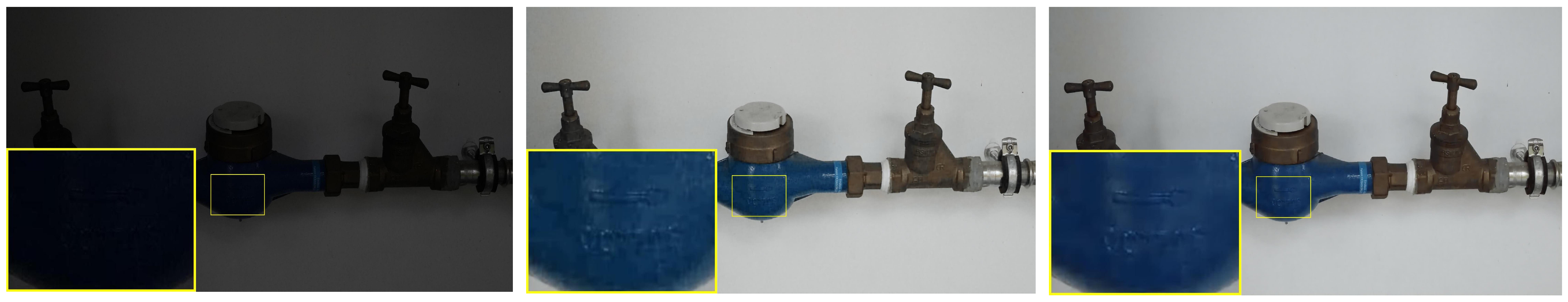} 
\\
\hspace{-4mm}\makebox[0.168\textwidth][c]{5.7896/0.4979}
\makebox[0.168\textwidth][c]{19.8100/0.9534}
\makebox[0.168\textwidth][c]{18.4409/0.9483}  
\\
\hspace{-4mm} \makebox[0.168\textwidth][c]{(a) Input}\makebox[0.168\textwidth][c]{ (b) $L=1$}
 \makebox[0.168\textwidth][c]{(c) $L=2$ }
\\
\hspace{-4mm}\includegraphics[width=0.49\textwidth]{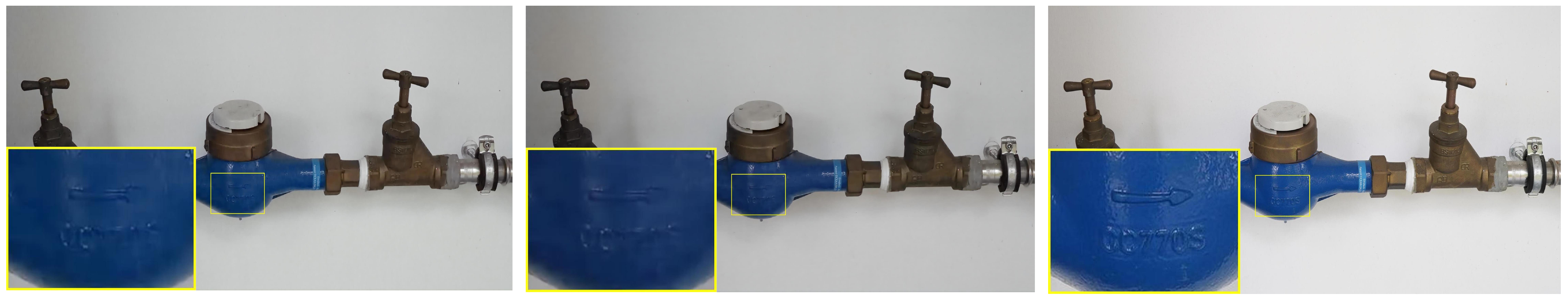}
\\
\hspace{-4mm}\makebox[0.168\textwidth][c]{\textbf{ 20.5723/0.9568} }
\makebox[0.168\textwidth][c]{18.3510/0.9482}\makebox[0.168\textwidth][c]{PSNR/SSIM}
\\
\hspace{-4mm} \makebox[0.168\textwidth][c]{\textbf{(d) $L=3$ (Ours)} }
 \makebox[0.168\textwidth][c]{(e) $L=4$}
 \makebox[0.168\textwidth][c]{(f) GT}
 \\
\end{tabular}
\caption{Visualization results on different numbers of PFI modules ($L$). Our model with $3$ PFI modules obtains results more desirable than others.}
\label{fig:Number of Prior Feature Interaction}
\end{center}
\end{figure}

\begin{figure*}[t]
\centering
\begin{center}
\begin{tabular}{ccccc}
\hspace{-4mm}\includegraphics[width=1\linewidth]{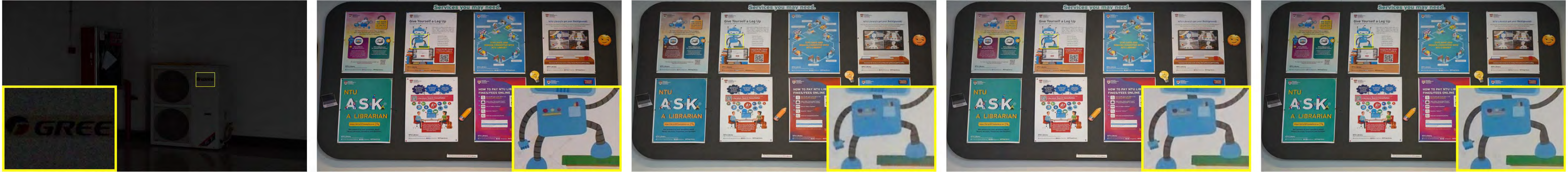} 
\\
 \makebox[0.195\textwidth][c]{13.2478/0.6616}  \makebox[0.195\textwidth][c]{PSNR/SSIM}
  \makebox[0.195\textwidth][c]{25.2194/0.9105}   \makebox[0.195\textwidth][c]{\textbf{25.2591}/0.9157}  \makebox[0.195\textwidth][c]{24.8555/\textbf{0.9241}}
\\
 \makebox[0.195\textwidth][c]{(a) Input}  \makebox[0.195\textwidth][c]{(b) GT}
  \makebox[0.195\textwidth][c]{(c) $C=8$}   \makebox[0.21\textwidth][c]{\textbf{(d) $C=16$ (Ours)}}  \makebox[0.195\textwidth][c]{(e) $C=32$}
\end{tabular}
\caption{Visualization results on different numbers of channels $C$. Our model with $16$ channels strikes a balance, producing results with enhanced texture details and more natural colors.}
\label{fig:Visualization results of the number of channel $C$.}
\end{center}
\end{figure*}

We present performance comparisons with state-of-the-art approaches on $5$ UHD image restoration tasks, including
\textbf{(a)} low-light enhancement, \textbf{(b)} dehazing, \textbf{(c)} deblurring, \textbf{(d)} desnowing, and \textbf{(e)} deraining.
%\vspace{-2mm}
\textbf{\subsection{Experimental Setup}}
\textbf{Implementation Details.} We incorporate $3$ PFI modules in low-resolution space for the network setting based on NAFBlocks (total $29$ in this paper) backbones, and $4$ NAFBlocks follow each PFI.
The number of attention heads in MTCA is set to $8$, and the number of channels is $16$.
We train models using AdamW optimizer with the initial learning rate $5e^{-4}$ gradually reduced to $1e^{-7}$ with the cosine annealing~\citep{loshchilov2016sgdr}.
The model is trained with a total batch size of $12$ on two NVIDIA RTX A6000 GPUs.
During training, we utilize cropped patches with a size of $512$$\times$$512$ for 60K iterations in low-light image enhancement, 500K iterations in desnowing, deraining, and dehazing, and 600K iterations in deblurring.
\\
\\
\noindent\textbf{Datasets.} We use the UHD-LL~\citep{LiGZLZFL23}, UHD-Blur~\citep{wang2024uhdformer}, and UHD-Haze~\citep{wang2024uhdformer} in line with previous works~\citep{LiGZLZFL23,wang2024uhdformer} as the UHD low-light image enhancement, deblurring, and dehazing benchmarks, respectively.
For UHD image desnowing and deraining, we use our proposed UHD-Snow and UHD-Rain datasets to evaluate the desnowing and deraining performance, respectively.
\begin{table*}[t]
\setlength{\tabcolsep}{17.5pt}
\caption{Low-light image enhancement on UHD-LL dataset. General and UHD respectively denote the general and UHD image restoration methods. The best and second best are marked in \textbf{bold} and \underline{underlined}, respectively.
$\uparrow$($\downarrow$) means higher(lower) is better.}
\label{tab:Low-light image enhancement.} 
\centering
\begin{tabular}{l|l|c|ccc}
\Xhline{1.5pt}
\textbf{Type}
&\textbf{Methods}
&\textbf{Venue}
&\textbf{~~PSNR~$\uparrow$~} 
&\textbf{SSIM~$\uparrow$~}
&\textbf{LPIPS~$\downarrow$~}
\\
\Xhline{1pt}
\multirow{3}{*}{\textbf{General}} 
&SwinIR~\citep{liang2021swinir}& ICCVW’21&21.165 & 0.8450 &0.3995 \\ 
&Restormer~\citep{Zamir2021Restormer}&CVPR’22&21.536 & 0.8437 &0.3608  \\ &Uformer~\citep{wang2021uformer}&CVPR’22&21.303   & 0.8233  &0.4013     \\
\Xhline{1pt}
\multirow{4}{*}{\textbf{UHD}} 
&LLFormer~\citep{LLformer}&AAAI’23&24.065 & 0.8580 &0.3516  \\
&UHDFour~\citep{LiGZLZFL23}&ICLR’23&26.226 & 0.9000 &0.2390   \\
&UHDformer~\citep{wang2024uhdformer}&AAAI’24&\textbf{27.113} & \underline{0.9271} &\underline{0.2240}   \\ 
&\textbf{UHDDIP (Ours)}&-&\underline{26.749}&\textbf{0.9281} &\textbf{0.2076}    \\
\Xhline{1.5pt}
\end{tabular}
\end{table*}
\begin{figure*}[t]
\centering
\begin{center}
\begin{tabular}{ccccccccc}
\hspace{-2mm}\includegraphics[width=1\linewidth]{images/lowlight_UHDLL.pdf}
\\
\end{tabular}
\caption{Low-light image enhancement on UHD-LL. Our UHDDIP can generate clearer results.
}
\label{fig:Low-light image enhancement on UHD-LL}
\end{center}
\end{figure*}
\begin{table*}[t]
\setlength{\tabcolsep}{17.5pt}
\caption{Image dehazing on the UHD-Haze dataset. The best and second best are marked in \textbf{bold} and \underline{underlined}.}
\label{tab:image dehazing.} 
\centering
\begin{tabular}{l|l|c|ccc}
\Xhline{1.5pt}
\textbf{Type}
&\textbf{Methods}
&\textbf{Venue}
&\textbf{~~PSNR~$\uparrow$~} 
&\textbf{SSIM~$\uparrow$~}
&\textbf{LPIPS~$\downarrow$~}
\\
\Xhline{1pt}
\multirow{3}{*}{\textbf{General}} 
&Restormer~\citep{Zamir2021Restormer}&CVPR’22&12.718  &0.6930  &0.4560    \\
&Uformer~\citep{wang2021uformer}&CVPR’22&19.828 &0.7374 &0.4220   \\
&DehazeFormer~\citep{DehazeFormer}&TIP’23&15.372 &0.7045 &0.3998 \\
\Xhline{1pt}
\multirow{3}{*}{\textbf{UHD}} 
&UHD~\citep{ZhengRCHWSJ21}&ICCV’21&18.048 &0.8113 &0.3593    \\
&UHDformer~\citep{wang2024uhdformer}&AAAI’24& \underline{22.586} &\underline{0.9427} &\underline{0.1188}   \\
&\textbf{UHDDIP (Ours)}&-&\textbf{24.699} &\textbf{0.9520} &\textbf{0.1049}   \\
\Xhline{1.5pt}
\end{tabular}
\end{table*}
\begin{figure*}[t]
\centering
\begin{center}
\begin{tabular}{ccccccccc}
\hspace{-2mm}\includegraphics[width=1\linewidth]{images/UHDdehaze.pdf}
\\
\end{tabular}
\caption{Image dehazing on UHD-Haze. Our UHDDIP is capable of producing clearer results.}
\label{fig:image dehazing on UHD-Haze}
\end{center}
\end{figure*}
%%%%%%%%%%%%%%%%%%%%%%%%%
\begin{table*}[t]
\setlength{\tabcolsep}{17.8pt}
\caption{Image deblurring on the UHD-Blur dataset. The best and second best are marked in \textbf{bold} and \underline{underlined}.}
\label{tab:image deblurring.} 
\centering
\begin{tabular}{l|l|c|ccc}
\Xhline{1.5pt}
\textbf{Type}
&\textbf{Methods}
&\textbf{Venue}
&\textbf{~~PSNR~$\uparrow$~} 
&\textbf{SSIM~$\uparrow$~}
&\textbf{LPIPS~$\downarrow$~}
\\
\Xhline{1pt}
\multirow{4}{*}{\textbf{General}} 
&Restormer~\citep{Zamir2021Restormer}&CVPR’22&25.210  &0.7522  &0.3695    \\
&Uformer~\citep{wang2021uformer}&CVPR’22&25.267 &0.7515 &0.3851   \\
&Stripformer~\citep{Stripformer}&ECCV’22&25.052 &0.7501 &0.3740 \\
&FFTformer~\citep{FFTformer}&CVPR’23&25.409 &0.7571 &0.3708 \\
\Xhline{1pt}
\multirow{2}{*}{\textbf{UHD}} 
&UHDformer~\citep{wang2024uhdformer}&AAAI’24& \underline{28.821} &\underline{0.8440} &\underline{0.2350}   \\
&\textbf{UHDDIP (Ours)}&-&\textbf{29.517} &\textbf{0.8585} &\textbf{0.2127}   \\
\Xhline{1.5pt}
\end{tabular}
\end{table*}
\begin{figure*}[!t]
\centering
\begin{center}
\begin{tabular}{ccccccccc}
\hspace{-2mm}\includegraphics[width=1\linewidth]{images/UHDdeblur.pdf}
\\
\end{tabular}
\caption{Image deblurring on UHD-Blur. Our UHDDIP is capable of producing clearer results.}
\label{fig:image deblurring on UHD-Blur}
\end{center}
\end{figure*}
%%%%%%%%%%%%%%%%%%%%%%%%%
\\
\noindent\textbf{Evaluation Metrics.} We quantitatively measure restored performance by reporting the PSNR~\citep{PSNR_thu}, SSIM~\citep{SSIM_wang}, and LPIPS~\citep{Zhang_lpips} of all compared methods.
Following~\citep{LiGZLZFL23, wang2024uhdformer}, for methods that cannot directly process full-resolution UHD images, we resize the input image to the maximum size the model can handle and return to the original size after testing.
\\
\\
\noindent\textbf{Compared Methods.} We adopt seven general IR methods such as SwinIR~\citep{liang2021swinir}, Uformer~\citep{wang2021uformer}, Restormer~\citep{Zamir2021Restormer}, DehazeFormer~\citep{DehazeFormer}, Stripformer~\citep{Stripformer}, FFTformer~\citep{FFTformer}, and SFNet~\citep{sfpnet}, and four UHD IR methods (i.e., LLFormer~\citep{LLformer}, UHDFour~\citep{LiGZLZFL23}, UHD~\citep{ZhengRCHWSJ21}, and UHDformer~\citep{wang2024uhdformer}) as benchmarks. For a fair comparison, we retrain the models according to the officially released codes of these methods and use the weights with the same number of iterations as our method for testing purposes.
\vspace{-2mm}
\textbf{\subsection{Main Results}}
\noindent\textbf{UHD Low-Light Image Enhancement.} In line with previous work ~\citep{LiGZLZFL23,wang2024uhdformer}, we evaluate the UHD low-light image enhancement on UHD-LL~\citep{LiGZLZFL23} as summarized in Table~\ref{tab:Low-light image enhancement.}.
Despite a slight inferiority in PSNR in comparison to UHDformer~\citep{wang2024uhdformer}, our method outperforms UHDformer in terms of SSIM and LPIPS. 
It is clearly witnessed that our UHDDIP can obtain excellent perceptual quality.
Fig.~\ref{fig:Low-light image enhancement on UHD-LL} presents a visual comparison, where UHDDIP can generate a clearer structure and more natural color. 
\\
\\
\noindent\textbf{UHD Image Dehazing.} Table~\ref{tab:image dehazing.} shows the evaluation of our UHDDIP and other dehazing methods on the UHD-Haze testing set. Our UHDDIP consistently outperforms general and UHD IR methods.
Compared to UHDformer~\citep{wang2024uhdformer}, our method achieves a substantial gain of $2.113$dB PSNR.
We provide visual examples in Fig.~\ref{fig:image dehazing on UHD-Haze} for UHD dehazing. It can be seen that our UHDDIP is effective in removing haze and restoring images that are visually closer to the ground truth than those of the other approaches. 
However, UHDformer~\citep{wang2024uhdformer} can tend to over-defog, resulting in a noticeable loss of detail.
%%%%%%%%%%%%%%%%%%%desnowing
\begin{table*}[t]
\setlength{\tabcolsep}{17.8pt}
\caption{Image desnowing on the UHD-Snow dataset. The best and second best are marked in \textbf{bold} and \underline{underlined}.}
\label{tab:image desnowing.} 
\centering
\begin{tabular}{l|l|c|ccc}
\Xhline{1.5pt}
\textbf{Type}
&\textbf{Methods}
&\textbf{Venue}
&\textbf{~~PSNR~$\uparrow$~} 
&\textbf{SSIM~$\uparrow$~}
&\textbf{LPIPS~$\downarrow$~}
\\
\Xhline{1pt}
\multirow{3}{*}{\textbf{General}} 
&Uformer~\citep{wang2021uformer}&CVPR’22&23.717 & 0.8711 &0.3095   \\
&Restormer~\citep{Zamir2021Restormer}&CVPR’22&24.142  & 0.8691  &0.3190      \\
&SFNet~\citep{0001TBRGC0K23sfnet}&ICLR’23&23.638 &0.8456 &0.3528  \\
\Xhline{1pt}
\multirow{3}{*}{\textbf{UHD}} 
&UHD~\citep{ZhengRCHWSJ21}&ICCV’21&29.294 &0.9497 &0.1416  \\
&UHDformer~\citep{wang2024uhdformer}&AAAI’24&\underline{36.614} &\underline{0.9881} &\underline{0.0245}    \\
&\textbf{UHDDIP (Ours)}&-&\textbf{41.563}&\textbf{0.9909} &\textbf{0.0179}     \\
\Xhline{1.5pt}
\end{tabular}
\end{table*}
\begin{figure*}[t]
\centering
\begin{center}
\begin{tabular}{ccccccccc}
\hspace{-2mm}\includegraphics[width=1\linewidth]{images/UHDdesnow.pdf}
\\
\end{tabular}
\caption{Image desnowing on UHD-Snow. Our UHDDIP is able to generate clearer results.}\label{fig:image desnowing on UHD-Snow}
\end{center}
\end{figure*}
%%%%%%%%%%%%%%%%%%
\\
\\
\noindent\textbf{UHD Image Deblurring.} 
Table~\ref{tab:image deblurring.} compares UHDDIP with the general IR method and UHD IR backbone networks on the UHD-Blur testing set, our UHDDIP demonstrates superior performance. For example, compared to UHD methods, UHDDIP is $0.696$dB in terms of PSNR higher than UHDformer~\citep{wang2024uhdformer}. In addition, UHDDIP outperforms the general methods by considerable margins in terms of PSNR, SSIM, and LPIPS.
A visual comparison of UHDDIP with other approaches is also illustrated in Fig.~\ref{fig:image deblurring on UHD-Blur}. Our method produces mostly clear visual results and is closer to the ground truth. 
\\
\\
\noindent\textbf{UHD Image Desnowing.} 
We implement UHD desnowing experiment on the constructed UHD-Snow dataset.
Table~\ref{tab:image desnowing.} summarizes the quantitative results.
One can obviously observe that UHDDIP achieves the best performance.
Our UHDDIP yields a $4.949$dB PSNR performance gain over the previous best method UHDformer~\citep{wang2024uhdformer}.
Fig.~\ref{fig:image desnowing on UHD-Snow} shows that UHDDIP effectively removes snow and produces a clearer structure while preserving more details.
\\
\\
\noindent\textbf{UHD Image Deraining.}
% \subsubsection{UHD Image Deraining}
We evaluate UHD image deraining with the constructed UHD-Rain dataset. 
The results are reported in Table~\ref{tab:image deraining.}.
As one can see UHDDIP significantly advances current state-of-the-art approaches. 
Especially, our UHDDIP achieves a substantial gain of $2.828$dB PSNR compared to UHDformer~\citep{wang2024uhdformer}.
Fig.~\ref{fig:image deraining on UHD-Rain} shows our UHDDIP effectively removes rain streaks and generates visually pleasant rain-free images, whereas existing methods often struggle to restore UHD images well.
%%%%%%%%%%%%%%%%%%%%%deraining
\begin{table*}[t]
\setlength{\tabcolsep}{17.5pt}
\caption{Image deraining on the UHD-Rain dataset. The best and second best are marked in \textbf{bold} and \underline{underlined}.}
\label{tab:image deraining.} 
\centering
\begin{tabular}{l|l|c|ccc}
\Xhline{1.5pt}
\textbf{Type}
&\textbf{Methods}
&\textbf{Venue}
&\textbf{~~PSNR~$\uparrow$~} 
&\textbf{SSIM~$\uparrow$~}
&\textbf{LPIPS~$\downarrow$~}
\\
\Xhline{1pt}
\multirow{3}{*}{\textbf{General}} 
&Uformer~\citep{wang2021uformer}&CVPR’22&19.494 &0.7163 &0.4598   \\
&Restormer~\citep{Zamir2021Restormer}&CVPR’22&19.408  &0.7105  &0.4775    \\
&SFNet~\citep{0001TBRGC0K23sfnet}&ICLR’23&20.091 &0.7092 &0.4768 \\
\Xhline{1pt}
\multirow{3}{*}{\textbf{UHD}} 
&UHD~\citep{ZhengRCHWSJ21}&ICCV’21&26.183 &0.8633 & 0.2885   \\
&UHDformer~\citep{wang2024uhdformer}&AAAI’24& \underline{37.348} &\underline{0.9748} &\underline{0.0554}   \\
&\textbf{UHDDIP (Ours)}&-&\textbf{40.176} &\textbf{0.9821} &\textbf{0.0300}   \\
\Xhline{1.5pt}
\end{tabular}
\end{table*}
\begin{figure*}[!t]
% \footnotesize
\centering
\begin{center}
\begin{tabular}{ccccccccc}
\hspace{-2mm}\includegraphics[width=1\linewidth]{images/UHDderain.pdf}
\\
\end{tabular}
\caption{Image deraining on UHD-Rain. Our UHDDIP is capable of producing clearer results.}
\label{fig:image deraining on UHD-Rain}
\end{center}
\end{figure*}
%%%%%%%%%%%%%%%%
\textbf{\subsection{Ablation Study}}
\noindent We conduct the ablation study to analyze the effect of each component on 3 UHD IR tasks including low-light image enhancement, deraining, and desnowing, and all models are trained on a patch of size $512 \times 512$ for 60K iterations.
\\
\\
\noindent\textbf{Effect of Prior Feature Interaction.} Since the prior feature interaction module plays one important role in our model, we investigate its effect on UHD low-light image enhancement, desnowing, and deraining in Table~\ref{tab:Ablation study}(a), (c), and (e), respectively.
\begin{table*}[t]\footnotesize
\caption{Ablation study. Each component in Prior Feature Interaction (PFI) is effective and each prior used in our model can enhance recovery quality.
}
\centering
\label{tab:Ablation study}
\subfloat[ Effect of PFI on UHD low-light image enhancement.
%SPFI and DPFI are utilized in conjunction to boost restoration performance.
\label{tab:Effect on FPI}
]
{
\begin{minipage}{0.467\linewidth}{\begin{center}
\setlength{\tabcolsep}{12.5pt}
\begin{tabular}{l|lccc}
\Xhline{1.5pt}
 \textbf{Module}& \textbf{PSNR}~$\uparrow$  &\textbf{SSIM~$\uparrow$} &\textbf{LPIPS~$\downarrow$} 
 %& \textbf{Parameters (M)~$\downarrow$}
\\
\Xhline{1pt}
w/o~PFI &26.499 &0.9269 &0.2169 \\
w/o~DPFI &26.226 &0.9232 &0.2179 \\
w/o~SPFI &26.557  &0.9259 &0.2171 \\
\textbf{Full (Ours)} &\textbf{26.749} &\textbf{0.9281} &\textbf{0.2076} \\
\Xhline{1.5pt}
\end{tabular}
% \end{tabular}
\end{center}}
\end{minipage}
}
\hspace{1.85em}
%#################################################
\subfloat[Effect of Prior on UHD low-light image enhancement. 
%Both normal and gradient priors improve restoration quality.
\label{tab:Effect on DPG}
]
{
\begin{minipage}{0.467\linewidth}{\begin{center}
\setlength{\tabcolsep}{12.5pt}
\begin{tabular}{l|lccc}
\Xhline{1.5pt}
 \textbf{Prior}& \textbf{PSNR}~$\uparrow$  &\textbf{SSIM~$\uparrow$} &\textbf{LPIPS~$\downarrow$} 
 %& \textbf{Running Time (s)}
\\
\Xhline{1pt}
w/o~Priors & 26.102 &0.9196 & 0.2405 \\
w/o~Gradient&26.002 &0.9253 &0.2137 \\
w/o~Normal &\textbf{26.975} &0.9272 &0.2096 \\
\textbf{Full (Ours)} &26.749 &\textbf{0.9281}&\textbf{0.2076} \\
\Xhline{1.5pt}
\end{tabular}
\end{center}}
\vspace{1mm}
\end{minipage}
}
\hspace{1.85em}
%#################################################
\subfloat[
Effect of PFI on UHD image desnowing.
\label{tab:Effect of PFI on desnowing.}
]
{
\begin{minipage}{0.467\linewidth}{\begin{center}
\setlength{\tabcolsep}{12.5pt}
\begin{tabular}{l|lccc}
\Xhline{1.5pt}
 \textbf{Module}& \textbf{PSNR}~$\uparrow$  &\textbf{SSIM~$\uparrow$} &\textbf{LPIPS~$\downarrow$} 
\\
\Xhline{1pt}
w/o~PFI &37.204 &0.9886 &0.0248 \\
w/o~DPFI &37.544 &0.9886 &0.0247 \\
w/o~SPFI &38.189  &0.9891 &0.0237 \\
\textbf{Full (Ours)} &\textbf{39.350} &\textbf{0.9894} &\textbf{0.0230} \\
\Xhline{1.5pt}
\end{tabular}
\end{center}}
\end{minipage}
}
\hspace{1.85em}
%##############################################
\subfloat[
Effect of Prior on UHD image desnowing. \label{tab:Effect of Prior on desnowing}
]
{
\begin{minipage}{0.467\linewidth}{\begin{center}
\setlength{\tabcolsep}{12.5pt}
\begin{tabular}{l|lccc}
\Xhline{1.5pt}
 \textbf{Prior}& \textbf{PSNR}~$\uparrow$  &\textbf{SSIM~$\uparrow$} &\textbf{LPIPS~$\downarrow$} 
\\
\Xhline{1pt}
w/o~Priors & 28.540 &0.9764 & 0.0583\\
w/o~Gradient&38.618 &0.9892 &0.0241\\
w/o~Normal &38.816 &0.9891 &0.0234 \\
\textbf{Full (Ours)} &\textbf{39.350} &\textbf{0.9894}&\textbf{0.0230} \\
\Xhline{1.5pt}
\end{tabular}
\end{center}}
\vspace{1mm}
\end{minipage}
}
\hspace{1.85em}
%##############################################
\subfloat[
Effect of PFI on UHD image deraining.
\label{tab:Effect of PFI on deraining.}
]
{
\begin{minipage}{0.467\linewidth}{\begin{center}
\setlength{\tabcolsep}{12.5pt}
\begin{tabular}{l|lccc}
\Xhline{1.5pt}
 \textbf{Module}& \textbf{PSNR}~$\uparrow$  &\textbf{SSIM~$\uparrow$} &\textbf{LPIPS~$\downarrow$} 
\\
\Xhline{1pt}
w/o~PFI &37.824 &0.9752 &0.0554 \\
w/o~DPFI &37.877&	0.9758&	0.0548 \\
w/o~SPFI &37.825  &0.9756&0.0535 \\
\textbf{Full (Ours)} &\textbf{38.072} &\textbf{0.9765} &\textbf{0.0522} \\
\Xhline{1.5pt}
\end{tabular}
\end{center}}
\end{minipage}
}
\hspace{1.85em}
%#################################################
\subfloat[
Effect of Prior on UHD image deraining. \label{tab:Effect of Prior on deraining}
]
{
\begin{minipage}{0.467\linewidth}{\begin{center}
\setlength{\tabcolsep}{12.5pt}
\begin{tabular}{l|lccc}
\Xhline{1.5pt}
 \textbf{Prior}& \textbf{PSNR}~$\uparrow$  &\textbf{SSIM~$\uparrow$} &\textbf{LPIPS~$\downarrow$} 
\\
\Xhline{1pt}
w/o~Priors & 33.096 &0.9530&0.1329  \\
w/o~Gradient& 37.966 & 0.9765& 0.0535 \\
w/o~Normal &38.005&0.9757&0.0546 \\
\textbf{Full (Ours)} &\textbf{38.072} &\textbf{0.9765} &\textbf{0.0522} \\
\Xhline{1.5pt}
\end{tabular}
\end{center}}
\end{minipage}
}
%#################################################
\label{tab:Ablation on Prompting Degradation Perception} 
\end{table*}
%%%%%%%%%%%%%%%%
\begin{figure*}[t]
\centering
\begin{center}
\begin{tabular}{ccccc}
\hspace{-2mm} \includegraphics[width=1\linewidth]{images/ab_pfi.pdf} 
\\
\makebox[0.195\textwidth][c]{(a) Input}  \makebox[0.195\textwidth][c]{(b) w/o PFI}
  \makebox[0.195\textwidth][c]{(c) w/o DPFI}   \makebox[0.195\textwidth][c]{(d) w/o SPFI}  \makebox[0.195\textwidth][c]{\textbf{(e) Full (Ours)}}
\end{tabular}
\caption{Visual comparison on each component in PFI  on UHD low-light image enhancement, deraining, and desnowing (from top to bottom).}
\label{fig:Visual comparison on PFI}
\end{center}
\end{figure*}
%%%%%%%%%%%%%%%%%%
We note that removing the PFI leads to a noticeable performance drop in all metrics on 3 tasks. 
For the UHD low-light image enhancement, only using the SPFI module degrades the network’s performance, while DPFI contributes to performance improvement. 
Whereas, utilizing in conjunction performs the best, and our method surpasses the baseline by $0.25$dB PSNR. 
Furthermore, it can be discovered that our SPFI and DPFI modules both play an important role in desnowing and deraining tasks.
Fig.~\ref{fig:Visual comparison on PFI} presents the visual comparison of 3 tasks, where our full model is able to generate results with clearer structures and more natural colors.
\\
\\
\noindent\textbf{Effect of Priors.} 
Table~\ref{tab:Ablation study}(b), (d), and (f) report the effect of the priors on 3 tasks, respectively.
We find that for UHD low-light image enhancement, the normal prior contributes to the improvement of SSIM and LPIPS compared without using any priors, while the gradient prior helps yield the optimal PSNR. 
When both, our method achieves the best performance in terms of SSIM and LPIPS.  
On the other hand, for desnowing and deraining, both normal and gradient priors contribute to improved restoration quality.
The analysis indicates that considering gradient and normal priors in model design may benefit tasks like desnowing/deraining that focus on preserving and enhancing details and structures. However, it may introduce noise in low-light image enhancement, resulting in lower PSNR and higher SSIM values.
Fig.~\ref{fig:Visual comparison on dual priors} illustrates a visual comparison on UHD low-light image enhancement. 
We observe that normal and gradient priors produce distinct structures and details (See (b) and (g) in Fig.~\ref{fig:Visual comparison on dual priors}). 
When combined, our method produces sharper structures and richer details.
\vspace{-2mm}
\textbf{\subsection{Additional Experiments}}\label{Additional experiments}
\noindent We conduct additional experiments to discuss the effect of each component used in our UHDDIP framework, including the effect on the number of prior feature interaction ($L$), the effect on the number of channels ($C$), the effect on the shuffle down factor ($S$), the effect on the shuffle down factor ($D$) in dual prior feature interaction, and the effect on loss functions. 
Finally, we provide the analysis for generalization on different resolutions.
Experiments are performed on the UHD-LL~\citep{LiGZLZFL23}, and models are trained on the patch of size $512 \times 512$ for 60K iterations.
\\
\\
\noindent\textbf{Effect on Number of Prior Feature Interaction ($L$).} 
In the prior feature interaction branch of UHDDIP, we analyze the impact on the number of prior feature interaction (PFI) modules for model performance.
Table~\ref{tab:Number of Prior Feature Interaction} demonstrates that performance incrementally improves as the number of PFI increases, alongside the number of parameters.
However, using $4$ PFI modules degrades the network’s performance, whereas $3$ PFI modules perform the best.  
We further present visual results, as illustrated in Fig.~\ref{fig:Number of Prior Feature Interaction}.
We note that a single PFI module often has difficulty removing the enhanced noise. 
Using $2$ PFI modules can effectively remove noise, but it challenges color restoration. 
The desirable results are produced using $3$ PFI modules. 
However, $4$ PFI modules will bring about a color difference, which may be because too many PFI modules make excessive interactions between prior features deteriorate the final result.
\begin{figure*}[t]
\centering
\begin{center}
\begin{tabular}{c}
\hspace{-2mm}\includegraphics[width=1\linewidth]{images/ab_priors.pdf}
\end{tabular}
\caption{Visual comparison of adopting different priors in the low-resolution branch on UHD low-light image enhancement.}
\label{fig:Visual comparison on dual priors}
\end{center}
\end{figure*}
\begin{table}[t]
\vspace{-3mm}
\setlength{\tabcolsep}{10.5pt}
\centering
\caption{Effect on Number of Prior Feature Interaction ($L$). The model with $3$ PFI modules achieves the best performance.}
\label{tab:Number of Prior Feature Interaction} 
\begin{tabular}{l|ccc}
\Xhline{1.5pt}
 \textbf{Number $L$ of PFI}& \textbf{PSNR}~$\uparrow$  &\textbf{SSIM~$\uparrow$} &\textbf{LPIPS~$\downarrow$}
\\
\Xhline{1pt}
$L=1$ & 25.902 &0.9227 &0.2336 \\
$L=2$ &26.352 &0.9254 &0.2162  \\
\textbf{$L=3$ (Ours)} &\textbf{26.749} &\textbf{0.9281}&\textbf{0.2076} \\
$L=4$ &26.479  &0.9269&0.2077\\
\Xhline{1.5pt}
\end{tabular}
\vspace{-3mm}
\end{table}
\\
\\
\noindent\textbf{Effect on Number of Channels ($C$).}
We measure the impact on the number of channels in all modules, as seen in Table~\ref{tab: Effect on number of channel.}.
It can be noticed that SSIM and LPIPS metrics improve as the number of channels increases. 
Our method with $16$ channels achieves optimal PSNR performance. 
However, increasing the number of channels to $32$ results in significant degradation of PSNR performance and increases the number of parameters by approximately $3.8$ times.
\begin{table}[t]
\setlength{\tabcolsep}{11.75pt}
\caption{Effect on Number of Channels ($C$).
The model with $16$ channels achieves the best performance in terms of PSNR.}
\label{tab: Effect on number of channel.} 
\centering
\begin{tabular}{l|ccc}
\Xhline{1.5pt}
 \textbf{Channels $C$}& \textbf{PSNR}~$\uparrow$  &\textbf{SSIM~$\uparrow$} &\textbf{LPIPS~$\downarrow$}
% & \textbf{Parameters (M)~$\downarrow$}
\\
\Xhline{1pt}
$C=8$ &26.602 &0.9244&0.2498  \\ 
\textbf{$C=16$ (Ours)} &\textbf{26.749} &0.9281&0.2076\\
$C=32$ &26.192  &\textbf{0.9283}&\textbf{0.1997}\\
\Xhline{1.5pt}
\end{tabular}
\end{table}
\begin{table}[t]
\setlength{\tabcolsep}{12.4pt}
\caption{Effect on Shuffle-Down Factor ($S$).
The model with the shuffle-down factor of $8$ is optimal.
}
\label{tab: Effect on Shuffle Down Factor.} 
\centering
\begin{tabular}{l|ccc}
\Xhline{1.5pt}
 \textbf{Factor $S$}& \textbf{PSNR}~$\uparrow$  &\textbf{SSIM~$\uparrow$} &\textbf{LPIPS~$\downarrow$}
 %& \textbf{Parameters (M)~$\downarrow$}
\\
\Xhline{1pt}
$S=4$ &26.415  &0.9259 &  0.2078\\ 
\textbf{$S=8$ (Ours)}  &\textbf{26.749} &\textbf{0.9281}&\textbf{0.2076}\\
$S=16$  &26.492  &0.9244 &0.2169  \\ 
\Xhline{1.5pt}
\end{tabular}
\end{table}
\begin{table}[t]
\setlength{\tabcolsep}{10.5pt}
\caption{Effect on Shuffle-Down Factor ($D$) in DPFI.
The model with the shuffle-down factor of $4$ obtains the best performance in terms of PSNR and SSIM.}
%\vspace{-2mm}
\label{tab: Effect on Shuffle Down Factor of DPFI} 
\centering
\begin{tabular}{l|ccc}
\Xhline{1.5pt}
 \textbf{Factor $D$ in DPFI}& \textbf{PSNR}~$\uparrow$  &\textbf{SSIM~$\uparrow$} &\textbf{LPIPS~$\downarrow$}
 %& \textbf{Parameters (M)~$\downarrow$}
\\
\Xhline{1pt}
$D=2$ &26.330  &0.9261 &\textbf{0.2061}	 \\	 	
\textbf{$D=4$ (Ours)}  &\textbf{26.749} &\textbf{0.9281}&0.2076\\
$D=8$&26.389  &0.9250 &0.2160	\\
\Xhline{1.5pt}
\end{tabular}
\end{table}
\begin{table}[t]
\setlength{\tabcolsep}{7.3pt}
\caption{Effect on Loss Functions. The model with total loss obtains the best performance.}\label{tab: Effect on loss function} 
\centering
\begin{tabular}{l|ccc}
\Xhline{1.5pt}
 \textbf{Loss Function}& \textbf{PSNR}~$\uparrow$  &\textbf{SSIM~$\uparrow$} &\textbf{LPIPS~$\downarrow$}
% & \textbf{Parameters (M)~$\downarrow$}
\\
\Xhline{1pt}
(a) w/o low-resolution loss& 26.072 &0.9248 &0.2128	\\	 	
\textbf{(b) Total loss (Ours)}  &\textbf{26.749} &\textbf{0.9281}&\textbf{0.2076} \\
\Xhline{1.5pt}
\end{tabular}
\end{table}
Thus, we finally set $C$ as 16 to make a trade-off between performance and computational cost.
In addition, we also provide a visual example in Fig.~\ref{fig:Visualization results of the number of channel $C$.}.
One can observe that when the channel is set as $8$, the recovered result still exhibits some noise and poor colors. 
Conversely, when the channel is set as $32$, the colors are better recovered, but they appear overly smooth, resulting in a loss of detailed textures.
In contrast, our model with $16$ channels strikes a balance, producing results with enhanced texture details and more natural colors.
\begin{figure*}[t]
\centering
\begin{center}
\begin{tabular}{ccccc}
\hspace{-4mm}\includegraphics[width=1\linewidth]{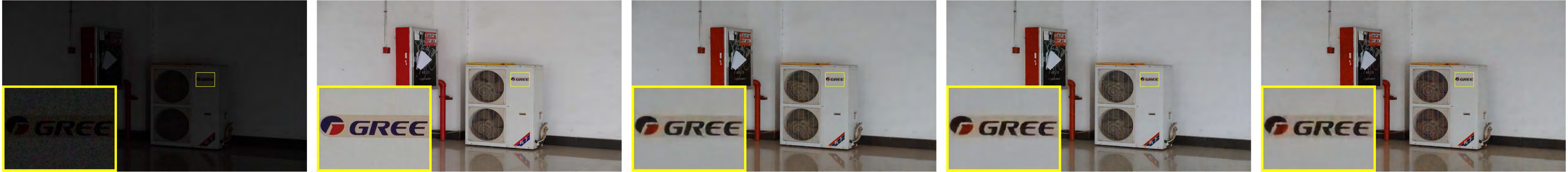} 
\\
 \makebox[0.195\textwidth][c]{8.1164/0.4579}
 \makebox[0.195\textwidth][c]{PSNR/SSIM}
 \makebox[0.195\textwidth][c]{26.7768/0.9543}
 \makebox[0.195\textwidth][c]{\textbf{30.6404/0.9539}}
 \makebox[0.195\textwidth][c]{28.2757/0.9517}
\\
 \makebox[0.195\textwidth][c]{(a) Input}
 \makebox[0.195\textwidth][c]{(b) GT }
 \makebox[0.195\textwidth][c]{(c) $S=4$ }
 \makebox[0.195\textwidth][c]{\textbf{(d) $S=8$ (Ours)}}
 \makebox[0.195\textwidth][c]{(e) $S=16$}
\end{tabular}
\caption{Visualization results on Shuffle-Down Factor $S$. Our model with the shuffle-down factor of $8$ is able to generate results with clearer structures and colors closer to GT.
}
\label{fig:Visual effect on the Shuffle Down Factor $S$}
\end{center}
\end{figure*}
\begin{figure*}[t]
\centering
\begin{center}
\begin{tabular}{ccccc}
\hspace{-2mm}\includegraphics[width=1\linewidth]{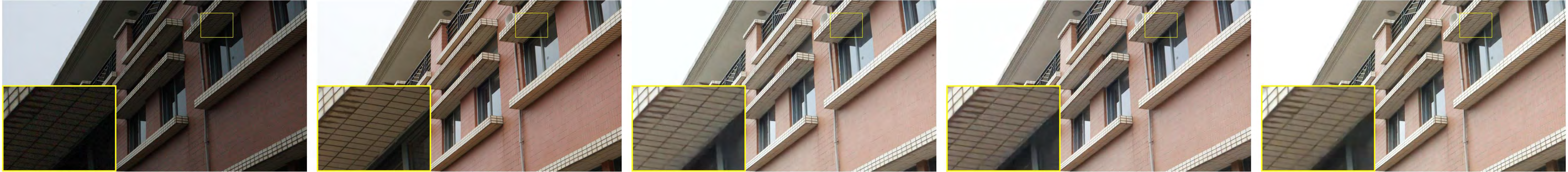}
\\
 \makebox[0.195\textwidth][c]{14.0047/0.6992}
 \makebox[0.195\textwidth][c]{PSNR/SSIM}
 \makebox[0.195\textwidth][c]{18.9252/0.8876}
 \makebox[0.195\textwidth][c]{\textbf{20.0564/0.9026}}
 \makebox[0.195\textwidth][c]{18.6056/0.8860}
\\
 \makebox[0.195\textwidth][c]{(a) Input}
 \makebox[0.195\textwidth][c]{(b) GT }
 \makebox[0.195\textwidth][c]{(c) $D=2$ }
 \makebox[0.195\textwidth][c]{\textbf{(d) $D=4$ (Ours)}}
 \makebox[0.195\textwidth][c]{(e) $D=8$}
\\
\end{tabular}
\caption{Visual effect on the Shuffle-Down Factor $D$ in DPFI. 
Our model with the shuffle-down factor of $4$ in DPFI is able to generate results with clearer structure and more details.
}
\label{fig:Visual effect on the Shuffle Down Factor $D$ in DPFI}
\end{center}
\end{figure*}
\begin{figure*}[!th]
\centering
\begin{center}
\begin{tabular}{cccc}
\hspace{-2mm}\includegraphics[width=1\linewidth]{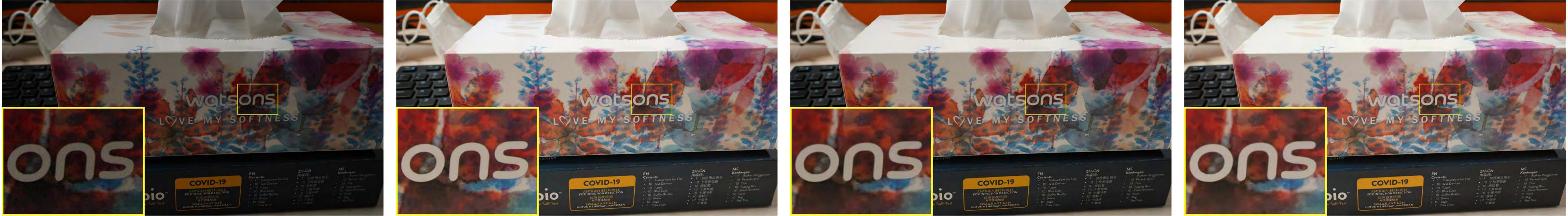} 
\\
\makebox[0.245\textwidth][c]{13.9304/0.7468}
\makebox[0.245\textwidth][c]{PSNR/SSIM }
\makebox[0.245\textwidth][c]{26.2085/0.9372 } \makebox[0.245\textwidth][c]{\textbf{36.5142/0.9448} }
\\
 \makebox[0.245\textwidth][c]{(a) Input}\makebox[0.245\textwidth][c]{ (b) GT}
 \makebox[0.245\textwidth][c]{(c) w/o low-resolution }
 \makebox[0.245\textwidth][c]{\textbf{(d) Total loss (Ours)} }
\\
\end{tabular}
\caption{Visualization results on different loss functions. Our model with total loss is able to generate a clearer structure and colors closer to GT.}
\label{fig:Effect on loss function}
\end{center}
\end{figure*}
\\
\\
\noindent\textbf{Effect on Shuffle-Down Factor ($S$).}
Table~\ref{tab: Effect on Shuffle Down Factor.} provides an ablation experiment on shuffle-down factor including $S=4$, $S=8$, and $S=16$. 
As can be seen, our model with the shuffle-down factor of $8$ is optimal. 
This suggests that using a resolution that is either too large or too small is detrimental to image recovery.
Fig.~\ref{fig:Visual effect on the Shuffle Down Factor $S$} shows a visual result. 
It is evident that the model with a shuffle-down factor of $8$ delivers superior visual quality compared to the other factor.
\begin{table*}[t]
\setlength{\tabcolsep}{18pt}
\caption{UHD low-light image enhancement by using the LOL training dataset~\citep{retinexnet_wei_bmvc18} and then testing on UHD-LL testing dataset. The best and second best are marked in \textbf{bold} and \underline{underlined}, respectively.}
\label{tab:Low-light image enhancement on LOL.} 
\centering
\begin{tabular}{l|l|c|ccc}
\Xhline{1.5pt}
\textbf{Type}
&\textbf{Methods}
&\textbf{Venue}
&\textbf{~~PSNR~$\uparrow$~} 
&\textbf{SSIM~$\uparrow$~}
&\textbf{LPIPS~$\downarrow$~}
\\
\Xhline{1pt}
\multirow{2}{*}{\textbf{General}}
&Restormer~\citep{Zamir2021Restormer}&CVPR’22&17.728 & 0.7703&0.4566  \\
&Uformer~\citep{wang2021uformer}&CVPR’22&18.168	&0.7201&0.5593 
\\
\Xhline{1pt}
\multirow{4}{*}{\textbf{UHD}} 
&LLFormer~\citep{LLformer}&AAAI’23&21.440	&0.7763&	0.4528
\\
&UHDFour~\citep{LiGZLZFL23}&ICLR’23&14.771	&0.3760&0.7608 
\\
&UHDformer~\citep{wang2024uhdformer}&AAAI’24&\textbf{22.615} & \underline{0.7754} &\underline{0.4241}   \\ 
&\textbf{UHDDIP (Ours)}&-&\underline{22.287}&\textbf{0.7898} &\textbf{0.4239}  \\
\Xhline{1.5pt}
\end{tabular}
\end{table*}
\begin{figure*}[t]
\centering
\begin{center}
\begin{tabular}{ccccccccc}
\hspace{-2mm}\includegraphics[width=1\linewidth]{images/lowlight_LOL.pdf}
\\
\end{tabular}
\caption{Visual results of UHD low-light image enhancement trained on the LOL dataset. Our UHDDIP can generate clearer results.
}
\label{fig:Low-light image enhancement trained on LOL}
\end{center}
\end{figure*}
\\
\\
\noindent\textbf{Effect on Shuffle-Down Factor ($D$) in Dual Prior Feature Interaction.}
To reduce the computational burden, we initially perform a shuffle-down operation on two prior features in DPFI. 
Table~\ref{tab: Effect on Shuffle Down Factor of DPFI} examines the impact of various shuffle-down factors on the model's performance. 
The results indicate that the network achieves optimal PSNR and SSIM with a shuffle-down factor of $4$.
A visual example is provided in Fig.~\ref{fig:Visual effect on the Shuffle Down Factor $D$ in DPFI}.
One can observe that, with a shuffle-down factor of $4$, the
the restored result on structures and details can be more desirable than others, which illustrates the effectiveness of this configuration.
%%%%%%%%%%%%%%%%%%%%%%%
\begin{table*}[t]
\centering
\setlength{\tabcolsep}{16.75pt}
\caption{Efficiency comparison. 
We report the number of parameters, FLOPs, and running time. 
The testing is conducted on a single RTX2080Ti GPU with a batch size of 1 at a resolution of $1024$$\times$$1024$.
}
\label{tab:Efficiency comparison} 
\begin{tabular}{l|c|ccc}
\Xhline{1.5pt}
 \textbf{Methods} & \textbf{Type}  &\textbf{Parameters (M)~$\downarrow$} &\textbf{FLOPs (G)~$\downarrow$} 
 & \textbf{Running Time (s)~$\downarrow$}
\\
\Xhline{1pt}
Restormer~\citep{Zamir2021Restormer}& \multirow{6}{*}{\textbf{General}}&26.10 &2255.85 & 1.86 \\
Uformer~\citep{wang2021uformer}&& 20.60 &657.45 &0.60\\
SFNet~\citep{0001TBRGC0K23sfnet}&&13.23 &1991.03 &0.61\\
DehazeFormer~\citep{DehazeFormer}&&2.51 &375.40 &0.45\\
Stripformer~\citep{Stripformer}&&19.71 &2728.08 &0.15\\
FFTformer~\citep{FFTformer}&&16.56 &2104.60 &1.27\\
\Xhline{1pt}
LLFormer~\citep{LLformer}& \multirow{5}{*}{\textbf{UHD}}&13.13 &221.64 & 1.69\\
UHD~\citep{ZhengRCHWSJ21}&&34.55  &113.45 &0.04 \\
UHDFour~\citep{LiGZLZFL23}&& 17.54  &75.63 &0.02  \\
UHDformer~\citep{wang2024uhdformer}&& \textbf{0.34}&48.37 &0.16\\
\textbf{UHDDIP (Ours)}& &0.81 &\textbf{34.73}  &0.13 \\
\Xhline{1.5pt}
\end{tabular}
\end{table*}
\\
\\
\noindent\textbf{Effect on Loss Functions.}\label{sec: Effect on Loss Function} In our method, we supervise not only the high-resolution space, but also the low-resolution space.
Hence, we are necessary to analyze the effect on loss functions.
Table~\ref{tab: Effect on loss function} shows that constraining both the high-resolution branch and the low-resolution branch together gains $0.677$dB PSNR gain compared to constraining only the high-resolution branch while having a similar number of parameters.
The finding suggests that significant enhancements can be achieved by applying an additional loss function to the low-resolution branch. 
This improvement likely stems from the greater utility of prior feature constraints in facilitating image recovery.
From Fig.~\ref{fig:Effect on loss function}, we can see that the model trained with the total loss function performs better, generating results with a clearer structure and colors closer to GT.
\\
\\
\noindent\textbf{Analysis for the generalization on different resolutions.} To deeper understanding of how models trained on low-resolution datasets generalize to UHD images, we conduct experiments on low-light image enhancement.
We retrain our model alongside several baseline methods using the low-resolution dataset LOL~\citep{retinexnet_wei_bmvc18} and test these models on UHD-LL. The results are reported in Table~\ref{tab:Low-light image enhancement on LOL.}.
As we can see, our UHDDIP performs outstandingly in SSIM and LPIPS scores, which is consistent with the results obtained by training on UHD-LL datasets and further illustrates that our model has generalizability on datasets of different resolutions.
Fig.~\ref{fig:Low-light image enhancement trained on LOL} provides visual results of UHD low-light image enhancement trained on the LOL dataset, where our UHDDIP is able to generate clearer results and visually closer to the ground truth.
\textbf{\subsection{Computational Complexity}}\label{sec: Computational Complexity}
\noindent Table~\ref{tab:Efficiency comparison} showcases the efficiency of various methods regarding the number of parameters, FLOPs, and running time.
These results are obtained on a single RTX2080Ti GPU, using a batch size of 1 at a resolution of $1024\times1024$. 
We observed that the proposed UHDDIP is effective, with significant improvements in parameters and FLOPs compared to the general IR methods including Uformer~\citep{wang2021uformer}, 
Restormer~\citep{Zamir2021Restormer}, 
SFNet~\citep{0001TBRGC0K23sfnet}, 
DehazeFormer~\citep{DehazeFormer},
Stripformer~\citep{Stripformer}, and
FFTformer~\citep{FFTformer},
and methods for a specific design for UHD restoration, e.g., LLFormer~\citep{LLformer}, UHDFour~\citep{LiGZLZFL23}, UHD~\citep{ZhengRCHWSJ21}, and UHDformer~\citep{wang2024uhdformer}.
Especially, our UHDDIP reduces by $28.2\%$ and $18.8\%$ in FLOPs and running time compared to UHDformer~\citep{wang2024uhdformer}, whereas it causes only a slight increase in parameters.
In conclusion, UHDDIP only increases a small number of parameters but significantly reduces computational complexity while achieving good performance.

\section{Conclusion}

In this paper, we have conducted new UHD benchmarks, including UHD-Snow and UHD-Rain to remedy the research on UHD image desnowing and deraining.
We have further proposed a dual interaction prior-driven UHD restoration method (UHDDIP) to address the problem of missing image details and structures in UHD restoration.
It is built around a prior feature interaction module containing SPFI and DPFI, that incorporates gradient and normal priors into model design to achieve high-quality restoration with finer structures and details.
Experimental results have shown that UHDDIP can achieve state-of-the-art performance on 5 tasks, including UHD low-light image enhancement, dehazing, deblurring, desonwing, and deraining.
%
% While priors’ spatial and detail benefit desnowing/deraining, it may lead to introducing noise in LLIE. 
% %
% In the future, we want to design a more general UHD image restoration model that can take into account noise introduction problems in LLIE conditions.
%\textbf{\textit{More analysis and visual results are provided in the supplementary material.}}
% \\
% \\
% \noindent\textbf{Acknowledgements} This research is partially supported by the National Key Research and Development Program of China (No.2018AAA0100300), and the National Natural Science Foundation of China (No.62306343).
\\
\\
\noindent\textbf{Data Availability Statement.} UHD datasets and pre-trained models used in this paper are available online. We provide corresponding source links for reproduction purposes in the \textbf{UHDDIP} repository \href{https://github.com/wlydlut/UHDDIP}{https://github.com/wlydlut/UHDDIP}.


{
% \small
%The spbasic bibliography style is designed to be used with natbib. So load natbib and then use \citet for Author (year) citations and \citep for (Author, year) citations.
\bibliographystyle{spbasic}
\bibliography{egbib}
}


%%%%%%%%%%%%%%%%%%%%%%%%%%%%%%%%%%%%%%%%%%%%%%%%%%%%%%%%%%%%%
% To-Do
% 1 change cite to citep citet
% 2 table mutil-row fix

% Related info
% 1 http://www.uco.es/~in1majim/calls/ijcv3dhumans.html

% Add related work
% 1 3d model based 
%     hmr? video-based smpl
%     VIBE: Video Inference for Human Body Pose and Shape Estimation
% 2 2d pose to 3d pose
%     3D human pose estimation in video with temporal convolutions and semi-supervised training

% Add experiments

\end{document}